\documentclass[lettersize,journal]{IEEEtran}
\usepackage{amsmath,amsfonts}
\usepackage{algorithmic}
\usepackage{algorithm}
\usepackage{array}
\usepackage[caption=false,font=normalsize,labelfont=sf,textfont=sf]{subfig}
\usepackage{textcomp}
\usepackage{stfloats}
\usepackage{url}
\usepackage{verbatim}
\usepackage{graphicx}
\usepackage{cite}
\usepackage{threeparttable}    

\usepackage{booktabs}
\usepackage{longtable}
\usepackage{hyperref}
\usepackage{textcomp}
\usepackage{upgreek}

\usepackage{xcolor}

\begin{document}
\title{When Vision Meets Touch: A Contemporary Review for Visuotactile Sensors from the Signal Processing Perspective}
\author{Shoujie Li, Zihan Wang, Changsheng Wu,  Xiang Li, Shan Luo, Bin Fang,
Fuchun Sun,~\IEEEmembership{Fellow,~IEEE,}  Xiao-Ping Zhang,~\IEEEmembership{Fellow,~IEEE,}
Wenbo Ding

\thanks{This work is supported in part by Shenzhen Key Laboratory of Ubiquitous Data Enabling (ZDSYS20220527171406015), Guangdong Innovative and Entrepreneurial Research Team Program (2021ZT09L197), Shenzhen Science and Technology Program (JCYJ20220530143013030),  Shenzhen Higher Education Stable Support Program  (WDZC20231129093657002), Tsinghua Shenzhen International Graduate School-Shenzhen Pengrui Young Faculty Program of Shenzhen Pengrui Foundation (No. SZPR2023005) and Meituan. (Corresponding author: Wenbo Ding). }
\thanks{Shoujie Li, Zihan Wang, Xiao-Ping Zhang, and Wenbo Ding are with Tsinghua Shenzhen International Graduate School, Shenzhen 518055, China. Wenbo Ding is also with the RISC-V International Open Source Laboratory, Shenzhen 518055, China. (e-mail: lsj20@mails.tsinghua.edu.cn;  zhwang22@mails.tsinghua.edu.cn; xpzhang@ieee.org; ding.wenbo@sz.tsinghua.edu.cn).}
\thanks{Changsheng Wu is with the Department of Materials Science and Engineering, National University of Singapore, Singapore 117575, Singapore. (e-mail:cwu@nus.edu.sg).  }
\thanks{Shan Luo is with the Centre for Robotics Research, King’s College London, London WC2R 2LS, U.K. (e-mail:shan.luo@kcl.ac.uk)}
\thanks{Xiang Li is with the Department of Automation, Tsinghua University, Beijing 100084, China. (e-mail: xiangli@tsinghua.edu.cn).}

\thanks{Bin Fang is with the School of Artificial Intelligence,
Beijing University of Posts and Telecommunications, Beijing 100876, China. (e-mail: fangbin1120@bupt.edu.cn).}

\thanks{ Fuchun Sun is with the Department of Computer Science and Technology, Tsinghua University, Beijing 100084, China. (e-mail: fcsun@mail.tsinghua.edu.cn).}

}

\maketitle
\begin{abstract}


Tactile sensors, which provide information about the physical properties of objects, are an essential component of robotic systems. The visuotactile sensing technology with the merits of high resolution and low cost has facilitated the development of robotics from environment exploration to dexterous operation. Over the years, several reviews on visuotactile sensors for robots have been presented, but few of them discussed the significance of signal processing methods to visuotactile sensors. Apart from ingenious hardware design, the full potential of the sensory system toward designated tasks can only be released with the appropriate signal processing methods. Therefore, this paper provides a comprehensive review of visuotactile sensors from the perspective of signal processing methods and outlooks possible future research directions for visuotactile sensors.
\end{abstract}


\section{Introduction}

\begin{figure}
	\centering
	\includegraphics[width=0.48\textwidth]{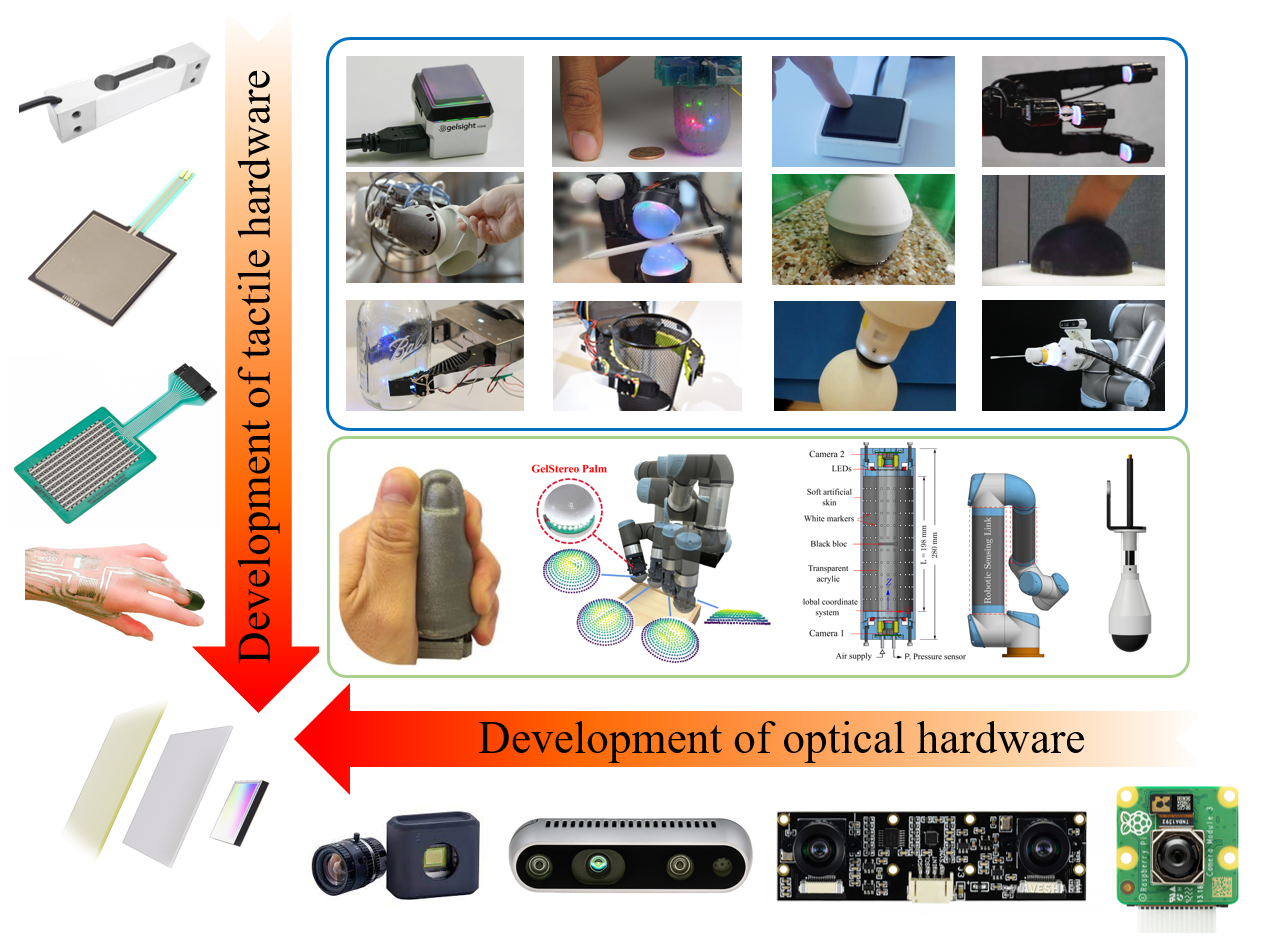}
	\caption{Trends in the intersections between the tactile sensor and optical hardware. Left: Tactile sensors from single-point force sensors to e-skin. Middle: The green boxes indicate the different parts of the robot designed using the visuotactile sensors, such as fingers~\cite{sun2022soft}, palms~\cite{cui2021hand}, arms~\cite{9247533}, and feet ~\cite{stone2020walking}. The blue box indicates the application of visuotactile sensors ~\cite{yuan2017GelSight,padmanabha2020omnitact,trueeb2020towards,lambeta2020digit,kuppuswamy2020soft,do2022densetact,li2022tata,kumagai2019event, liu2022GelSight,she2020exoskeleton, lepora2021pose,ma2019dense}.
 Bottom: Optical hardware from RGB cameras to event-based dynamic vision sensors. } \label{fig:1}
\end{figure}

\IEEEPARstart{W}{ith}   the rapid advancement of artificial intelligence, robots have increasingly been utilized for more intricate and complex tasks, such as industrial assembly~\cite{wan2017teaching,kyrarini2019robot}, human-robot collaboration, and surgery ~\cite{dupont2021decade,saeidi2022autonomous}. To perform these tasks, the robot must not only acquire the force in contact between the actuator and the environment but also the position of the end tool within the hand, which heavily relies on the resolution and accuracy of the tactile sensors. To improve the tactile perception of robots, tremendous sensors have been designed based on different mechanisms, such as piezoelectric~\cite{lin2021skin,wang2020energy}, triboelectric~\cite{tao2019self,song2022flexible}, and piezoresistive~\cite{yue2017piezoresistive,pei2021fully,rasouli2018extreme} sensors. Nevertheless, these sensors are limited by the complicated fabrication process and the expensive data acquisition circuits, and it is challenging to achieve high-resolution and large-scale tactile perception in a cost-efficient way.

Compared with tactile perception, visual perception by the external camera generally has a larger detection area. However, it is difficult to obtain the pose of the occluded object as well as the contact information during the manipulation. As shown in Fig.~\ref{fig:1}, with the advancement of optical imaging techniques, researchers have combined visual perception with tactile perception, which uses cameras to detect the deformation of the sensor surface~\cite{kakani2021vision}. Based on this mechanism, various genius visuotactile sensors have been designed, such as fingertip tactile sensors like GelSight~\cite{yuan2017GelSight}, Digit~\cite{lambeta2020digit}, robotic arms~\cite{asahina2019development,zhang2020vtacarm}, and robot feet~\cite{zhang2021tactile}. The most significant function of visuotactile sensors is 3-dimensional (3D) reconstruction. By utilizing high-resolution optical imaging methods, real-time reconstruction of the contact surfaces' 3D shape can be realized by photometric stereo~\cite{yuan2017GelSight,wang2021GelSight} and binocular imaging~\cite{zhang2018robot,cui2021hand} principles. In addition, the visuotactile sensor can also achieve contact area segmentation, high-resolution force perception ~\cite{obinata2007vision,sato2008measurement,ma2019dense}, slip detection~\cite{watanabe2008grip,yuan2015measurement,dong2017improved, Pneumatic}, and mapping and localization~\cite{chaudhury2022using,kuppuswamy2019fast,anzai2020deep}, which significantly improve the stability of object grasping and manipulation. Furthermore, visuotactile sensors can enable robots with more challenging tasks such as texture classification~\cite{yuan2017connecting,yuan2018active}, hardness classification~\cite{yuan2017shape}, underwater grasping~\cite{li2022tata}, cable manipulation~\cite{she2021cable}, etc.


While previous reviews on visuotactile sensors \cite{abad2020visuotactile,shimonomura2019tactile,zhang2022hardware,shah2021design} have discussed the sensor design and fabrication process, the role of signal processing in visuotactile sensors is rarely touched. Consequently, this paper summarizes the signal processing methods and applications of visuotactile sensors from the following new perspectives:

\begin{itemize}
\item The advantages and drawbacks of visuotactile systems with different structures in terms of sensing skin, illumination system, and vision system.
\item The signal processing techniques used in visuotactile sensors with respect to their performance in contact area segmentation, reconstruction, force perception, slip detection, mapping and localization, and simulation-to-reality (sim-to-real).
\item The applications, limitations, and the future development directions of visuotactile sensors.
\end{itemize}

The rest of the paper is organized as follows:
The sensor design of the visuotactile sensor is introduced in Section II.
The signal processing method of the visuotactile sensor is introduced in Section III.
Section IV presents the relevant applications of visuotactile sensors.
Section V discusses the current problems and future research directions of visuotactile sensors.
Finally, Section VI concludes this paper.

\section{Sensor Design}

\begin{figure}
	\centering
	\includegraphics[width=0.47\textwidth]{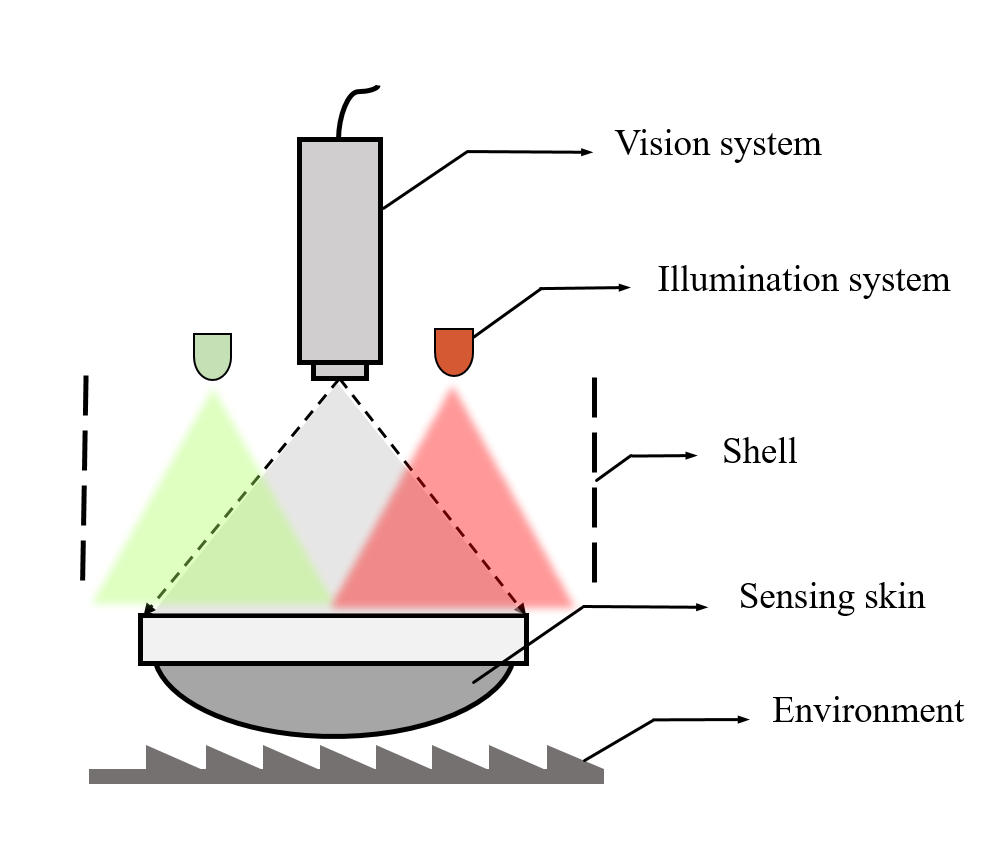}
	\caption{The structure of the visuotactile sensor, which includes sensing skin, illumination system, and vision system.} \label{fig:2}
\end{figure}

The structure of the visuotactile sensor can be divided into three parts: sensing skin, illumination system, and vision system, as shown in Fig.~\ref{fig:2}. The sensing skin is the core component of the visuotactile sensor, capable of detecting and representing information such as force, temperature, and texture through deformation or color changes upon contact with an object. The illumination system is tailored to the properties and functions of the sensing skin, enhancing the 3D geometric representation of the sensor. The vision system serves as the signal collection unit, capturing the deformation and color information generated by the sensing skin through optical imaging. The structure of the visuotactile sensor determines its functionality, and researchers have designed sensors of various parameters and sizes to meet different application scenarios.
A comprehensive summary of the mainstream visuotactile sensors is shown in Table.~\ref{tab1}. 

\subsection{Sensing skin}

\begin{figure}
	\centering
	\includegraphics[width=0.45\textwidth]{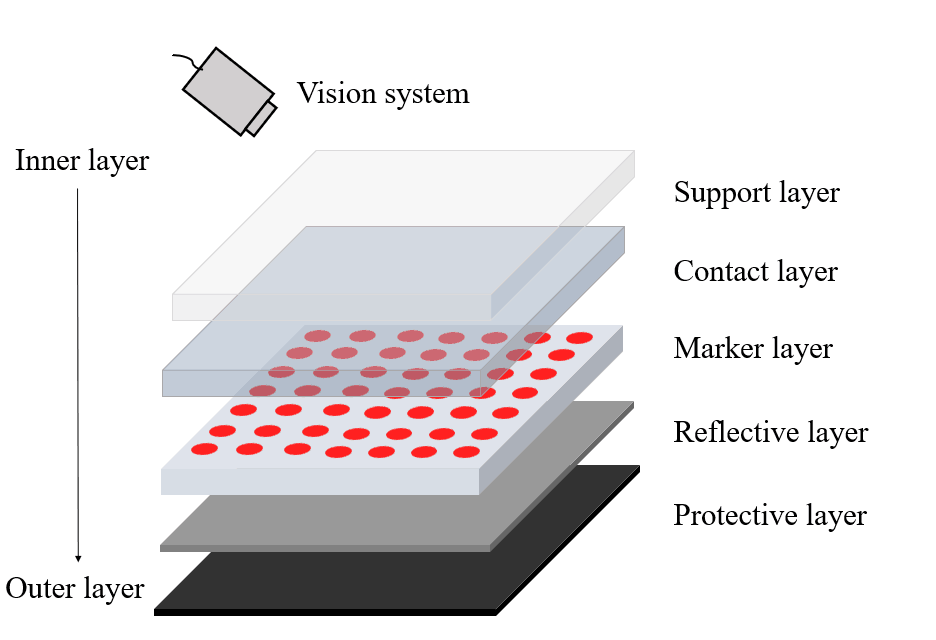}
	\caption{A typical structure of the sensing skin includes a support layer, contact layer, marker layer, reflective layer, and protective layer. } \label{fig:3}
\end{figure}

To capture more detailed texture and deformation information, sensing skins often adopt a multi-layer structure, which typically includes a protective layer, a reflective layer, a marker layer, a contact layer, and a support layer, as illustrated in Fig.~\ref{fig:3}. However, depending on the specific application, not all of these layers may be necessary. In the following sections, we will discuss the advantages and disadvantages of different sensing skin designs, taking into account factors such as shape, material, and markers.

\begin{table*}
\caption{Mainstream visuotactile sensor structural design and function}
\begin{center}
\begin{threeparttable}          

\begin{tabular}{m{2.6cm}<{\centering}|m{0.8cm}<{\centering}|m{1.9cm}<{\centering}|m{1.5cm}<{\centering}|m{1.5cm}<{\centering}|m{1.5cm}<{\centering}|m{1.8cm}<{\centering}|m{1.3cm}<{\centering}|m{1.3cm}<{\centering}}
\hline
Related works & Shape & Marker & Material & Illumination & Vision  &  Rconstruction & Force & Silp   \\\hline

GelForce~\cite{sato2009finger} &  3D       &  Double layer markers     & Silicone    &  White     & Monocular  & -  & FEM & -   \\
GelSight~\cite{yuan2017GelSight} & 2D &  Array markers &  Silicone & RGB & Monocular & PS & NN & CM  \\
TacTip~\cite{ward2018tactip} &  3D       &  Array markers     & Silicone    &  White     & Monocular  & -  &  & NN  \\
FingerVision~\cite{zhang2018fingervision} &  2D  & Array markers  & Silicone     &  White     & Monocular  & -  & FEM &   -  \\
Sferrazza  \textit{et al.}~\cite{sferrazza2019design}  &  2D  & Dense markers   & Silicone      &  White     & Monocular  & -  & FEM &  -   \\
Naeini \textit{et al.}~\cite{naeini2019novel}& 2D & -    &  Silicone    &  Red     &  Monocular    & - & NN  &  - \\

Kumagai  \textit{et al.}~\cite{kumagai2019event} & 3D & -   &  Elastomer gel    &  White     &  DVS    & -  &  - &  - \\
Lin  \textit{et al.}~\cite{lin2019sensing}  & 2D & Double layer markers    &  Silicone      &  White    &  Monocular   & -    &   RM &  -  \\
NeuroTac~\cite{ward2020miniaturised}  & 2D &  Array markers    & Silicone   &  White  &  DVS    & -  & -  & - \\
F-TOUCH~\cite{li2020f} & 2D & Array markers  &  Silicone     &  RGB     & Monocular   &   & RM &  -   \\
 Abad  \textit{et al.}~\cite{abad2020low} & 2D & UV markers   & Silicone   &  RGB +UV    &  Monocular    & -  &  Optical flow  &  Optical flow \\
Soft-bubble~\cite{kuppuswamy2020soft} &  3D     &  Dense  markers      & Latex   & IR   & Depth+IR  & ToF  & - &  -     \\ 
Digit~\cite{lambeta2020digit}   &    2D     &  -   & Silicone   & RGB  & Monocular  &- & -&  -     \\ 
OmniTact~\cite{padmanabha2020omnitact} & 3D & Dense pixels  &  Silicone       &  White    &  Camera array   & -  & - &  -   \\
Ouyang  \textit{et al.}~\cite{ouyang2020low} & 2D & Fiducial Tags  &  Resin       &  White    &  Monocular  & -  & RM & -    \\
GelFlex~\cite{she2020exoskeleton} &  3D & Array markers  &  Silicone     &  White     & Binocular   & NN  & - &  -   \\

NeuroTac~\cite{ward2020neurotac}& 3D & Array markers    &  Silicone    &  White     &  DVS    & - & - &  - \\
GelTip~\cite{gomes2020geltip}&    3D     &  -   & Silicone   & RGB  & Monocular  & PS  & -&    -   \\ 
Romero \textit{et al.}~\cite{romero2020soft}   & 3D & -  &  Silicone       &  White    &  Monocular   & PS  & - &  -   \\
Chen \textit{et al.}~\cite{yu2020vision}& 2D & -    &  TLC  ink    &  White    &  Monocular    & - & -  &  - \\
GelSight Wedge~\cite{wang2021GelSight} & 2D & Array markers  &  Silicone     &  RGB     & Monocular   & PS+NN  & - & -    \\
FingerVision with Whiskers~\cite{yamaguchi2021fingervision} & 2D & Whiskers &  Silicone     &  RGB     & Monocular   &   & RM &   -  \\
GelStereo Palm~\cite{cui2021hand} & 3D & Array markers  &  Silicone     &  White     & Binocular   & Binocular  & - &  -   \\
Viko~\cite{pang2021viko} & 2D & Dense pixels  &  Gecko patch       &  White    &  Monocular   &  -   & RM &  -   \\
Li \textit{et al.}~\cite{li2021design} & 3D & -    &  Latex      &  IR    &  Depth+
Monocular    & Structured light & -  &  - \\
Visiflex~\cite{fernandez2021visiflex} & 3D & LED markers  &  Silicone      &  RB    &  Monocular   &  -   & Mechanics model & -   \\
HaptiTemp~\cite{abad2021haptitemp} & 2D & UV markers  &  TM     &  UV and white     & Monocular   & Binocular  & - &  -   \\
InSight~\cite{sun2022soft}  &    3D     &  -   & Silicone   & RGB  & Monocular  & PS  & NN &-       \\ 
Gelsim~\cite{taylor2022gelslim}   & 2D        &  Array markers  & Silicone & RGB & Monocular  & PS  & FEM  & -       \\  
DenseTact~\cite{do2022densetact} &  3D     &  -   & Silicone   & RGB  & Monocular  & PS  & NN & -      \\ 
DTact~\cite{lin2022dtact} &  2D & -  &  Silicone     &  White     & Monocular  & Luminance  & - &  -   \\
DigiTac~\cite{lepora2022digitac} &  2D & Array markers  & Silicone     &  RGB      & Monocular  & -  & - &  -   \\

Soft-Jig~\cite{sakuma2022soft}  & 3D & -  &  Silicone       & -     &  Monocular   & -  & - &  -   \\
TacRot~\cite{zhang2022tacrot}  & 2D & -  &  Silicone       &  White    &  Monocular   &  PS   & - &   -  \\
TaTa~\cite{li2022tata}  & 2D & -  &  latex       &  RGB    &  Monocular   &  PS   & - & -    \\
Trueeb \textit{et al.}~\cite{trueeb2020towards} & 2D & -  &  Silicone       &  White    &  Camera array   &  -   & NN &  -   \\
GelSight Fin Ray~\cite{liu2022GelSight} & 3D & Array markers  &  Silicone     &  RGB     & Monocular   & PS  & - &   -  \\
Tac3D~\cite{zhang2022tac3d} &  3D     &  Array markers     & Silicone   & White   & Virtual binocular  & PS  & FEM &  CM      \\ 
Zhang \textit{et al.}~\cite{zhang2022multidimensional} & 2D & Dense pixels  &  Silicone       &  White    &  CMOS with pinhole   & -  & RM & -    \\
Faris  \textit{et al.}~\cite{faris2022proprioception} & 3D & Array markers   &  Silicone    &  White     &  DVS    & NN & - &  NN \\
UVtac~\cite{kim2022uvtac} & 2D & UV markers   &  Silicone    &  White     &  RGB    & -&  RM &  - \\

DelTact~\cite{zhang2022deltact} &  2D  & Dense pixels  & Silicone     &  White     & Monocular  & Optical flow  & FEM & NN    \\

Finger-STS~\cite{hogan2022finger}& 2D & UV markers    &  Silicone      &  UV    &  Monocular   &  -   &  -  &  CM  \\
Li \textit{et al.}~\cite{liimplementing} & 2D & ArUco markers  &  Silicone resin      &  -    & Monocular   & -  & - &  -   \\

FVSight~\cite{xiong2022fvsight }  & 2D & -   & Flexible  force sensor   &  RGB     &  Monocular    & -  &  Force tactile layer  &  - \\
SpecTac~\cite{wang2022spectac} & 2D & UV markers    &  Silicone      &  UV    &  Monocular   & -    & NN   &  -  \\
GelStereo~\cite{hu2023gelstereo}  &  2D       &  Array markers     & Silicone    &  White     & Binocular  & Binocular  & - & -      \\ 
DotView~\cite{zheng2023dotview} &  2D & Protrusions  & CS     &  -      & Capacitive sensor  & -  & NN &  -   \\
StereoTac~\cite{roberge2023stereotac} & 2D & -   &  Silicone    &  RB     &  Binocular    & PS + Binocular  &  - &  - \\

Althoefer  \textit{et al.}~\cite{althoefer2023miniaturised} & 3D &  Array markers    & Silicone   &  RGB  &  Monocular    & -  &  RM  & - \\

\hline

\end{tabular}

\begin{tablenotes}    
\footnotesize               
\item{Abbreviations:} Conductive silicone, CS; Time of flight, ToF; Thermosensitive materials, TM; Photometric stereo, PS; thermochromic liquid crystals, TLC;  Event-based dynamic vision sensor, DVS; Photometric stereo, PS; Ultraviolet, UV; Neural Networks, NN; The finite element method, FEM; Regression model, RM; Contact model, CM; Infrared, IR.
\end{tablenotes}            
\end{threeparttable}       
\end{center}
\label{tab1}
\end{table*}





\subsubsection{Shape}

Based on the sensor's surface geometry, the sensing skin can be categorized into two main types: 2D and 3D. In this paper, we define the visuotactile sensor with a marginal convex surface as 2D as well. As shown in Table~\ref{tab1}, 2D visuotactile sensors include GelSight~\cite{yuan2017GelSight}, Digit~\cite{lambeta2020digit}, etc., and 3D visuotactile sensors include GelTip \cite{gomes2020geltip,gomes2020blocks}, TouchRoller \cite{cao2023touchroller}, Soft-bubble~\cite{alspach2019soft}, Insight~\cite{sun2022soft}, TaTa~\cite{li2022tata}, etc. The 2D visuotactile sensor is typically mounted on the fingertip to sense geometry on a 2D plane. On the one hand, the 3D sensor is designed to be more versatile and can be mounted on fingertips or palms with an appropriate size, allowing it to sense the shape of the object from different angles. On the other hand, the 3D convex structure of the sensor not only enhances stability when grasping objects but also provides a larger sensing range. However, the 3D structure also has some problems, such as:

\begin{itemize}
\item Complex production process. Creating sensing skins with a 3D structure is difficult. Especially coating reflective or marker layers on the 3D surface with a high level of uniformity and durability.
\item Difficult signal processing. Image signal acquisition for 3D structure sensors can be a challenging task. One of the main difficulties is ensuring uniform illumination of the structure from all directions. Additionally, the magnitude and direction of the forces causing deformation at different contact points can vary greatly, making reconstruction and perception complex. In contrast, the signal processing for 2D structure is comparatively easier as it allows for better control of lighting and modeling, and the deformation is more consistent.
\item Challenges in calibration. Before conducting force detection, sensor calibration is often necessary, particularly for 3D structures where additional contact data must be gathered.
\end{itemize}

Most of the 3D tactile sensors currently available have a convex structure. However, Li \textit{et al.}  proposed a novel visuotactile sensor called CoTac~\cite{Pneumatic}, which has a concave design and is capable of sensing small tangential forces. This innovative sensor can be used in a variety of applications, including pharyngeal swab sampling and feeding.

\subsubsection{Marker}

The human hand possesses an exceptional tactile perception due to the abundance of sensory nerves on the skin's surface~\cite{saal2017simulating}. Inspired by this, researchers have enhanced the perception ability of visuotactile sensors by incorporating markers. Based on the size changes and displacement of the markers, the sensors can obtain normal force, tangential force, and slip signals.

In early works, the visuotactile sensor's markers were predominantly in a single color~\cite{ward2018tactip}, which makes it challenging to distinguish between tangential and normal forces. However, subsequent research focused on optimizing the marker's design to enhance its sensing capabilities. Katsunari \textit{et al.} proposed a two-layered structure with red and blue markers at the upper and bottom, respectively, to achieve more accurate force detection on convex surfaces~\cite{sato2008measurement}. By observing the relative offset of the markers, the magnitude and direction of the contact force can be obtained. Lin \textit{et al.} further optimized this structure by designing an array of diffusive and transmissive markers on the surface of the sensor, which is a square color array made of red and magenta markers~\cite{lin2019sensing}. 

Although  the markers can enhance the sensor's ability to detect force, they are sparse and cannot provide a high-resolution force distribution. To solve this problem, Zhang \textit{et al.}  utilized a dense color pattern instead of a dot matrix, which is a texture template composed of random pixel dots~\cite{zhang2022deltact}. This approach enables the sensor to capture denser point clouds and contact force information. Although increasing the density of markers on the sensor surface can improve force resolution, it reduces the ability to detect texture. Moreover, fabricating denser marker layers with consistent dot sizes and robustness to shift or detach during usage becomes difficult.

Despite existing problems, the design of the markers offers valuable insights for addressing force perception and slip detection. To mitigate the impact of markers on texture detection, a dual-modality switching method has been proposed~\cite{hogan2022finger}. This method uses ultraviolet (UV) fluorescent paint to create markers that are only visible under UV light, allowing for markers and texture detection to be achieved by switching between UV and white light.

\subsubsection{Material}

The detection effect of the visuotactile sensor can be influenced by the hardness, thickness, and transparency of the material used. Latex, silicone, and polydimethylsiloxane (PDMS) are commonly used materials for sensor skin. The choice of material is related to the application scenario and function of the sensor.

Silicone is widely used in the design of skin sensing for visuotactile sensors. It is a versatile material that can be used in the contact layer, marker layer, and reflective layer. Commercial silicone suppliers include Smooth-On, Silicones Inc, and WACKER. The advantages of silicone are as follows:

\begin{itemize}
\item Easy to mold. Simple to process, with minimal equipment requirements, and the model can be easily obtained through molding.
\item Design flexibility. By selecting various silicone materials or adjusting the mixing ratios, it is possible to achieve silicone materials with varying degrees of hardness and transparency.
\item Good compatibility. By incorporating diverse materials into silicone, such as silver powder and dye, it is possible to tailor the optical properties of coatings to meet specific requirements.
\end{itemize}

Compared with silicone, the process of producing latex film is more complex. Therefore, most sensors that use latex film~\cite{li2021design,kuppuswamy2020soft} are made from commercially available latex film, which is inexpensive but difficult to customize. Additionally, latex films are softer, which means that when gripping objects, the gas must often be injected into the sensor to increase its stiffness. Latex has the advantage of possessing higher elasticity and toughness, allowing it to conform better to the shape of the object being detected. As a result, it is frequently utilized as a sensing skin for larger visuotactile sensors.

Besides silicone and latex, PDMS can also serve as a material for creating visuotactile sensors. PDMS is known for its high transparency, but it is also harder than the other two materials. As a result, it is often utilized for sensor surfaces that require exceptional transparency~\cite{zhang2023improving}.

\subsubsection{Functional layers}

In addition to using reflective coating, researchers enriched the sensory functionality and improved the gripper's performance by introducing functional materials and structures in addition to using the reflective coating. 

In terms of more functions,  Fang \textit{et al.} developed a novel visuotactile sensor that incorporates both texture and temperature detection~\cite{fang2021novel}. The sensor is designed with a temperature-sensing region composed of three thermochromic materials, which can detect temperatures ranging from 5$^{\circ}$C to 45$^{\circ}$C. Hogan \textit{et al.} proposed a dual-mode semitransparent skin that can rapidly switch between the tactile sensor and visual camera mode by controlling the internal lighting conditions~\cite{hogan2021seeing}. The fusion of tactile and visual information provides a more effective approach to quantifying the physical properties of objects. The unique structural design of the sensor also enhances the gripping ability of the gripper. For grasping performance enhancement,  Pang \textit{et al.}  designed a soft gripper with a gecko palm-inspired self-adhesive layer~\cite{pang2021viko}. The layer with a micro-wedges structure significantly increases the grippers' load in handling objects with smooth surfaces.

In addition, the signal processing method of the visuotactile sensor often has to match the sensing skin. For example, for the sensing skin with markers, we often determine the contact force and information such as whether sliding occurs based on the displacement of the makers. While the perceptual skin without markers is more suitable for the deep learning method.

\subsection{Illumination system}

The design of the illumination system is determined by the structure of the sensing skin. To achieve different detection effects, people have to design special illumination circuits to match the sensing skin. Next, we will introduce two aspects of the illumination system: the installation position and the color.

The illumination system is mainly installed at the side and below the sensing skin. The design installed below the sensing skin has a larger illumination range, but this light is not directional and it is difficult to ensure the consistency of light intensity at each pixel point in a small space. And the method installed on the side of the sensing skin uses the sensing skin as a light waveguide, and the light will propagate inside the sensing skin. When the sensor is in contact with an object, the different directions of the contact position will show different colors. Based on this principle, the reconstruction of the contact area can be achieved.

There are mainly two types of light, white and RGB. The function of the white light is to improve the brightness inside the sensor, because the sensor is usually a closed structure, the brightness is very low in the absence of light. The RGB light can improve both the brightness inside the sensor and the contrast of the perceived skin surface pattern and make it directional. In addition, Hogan \textit{et al.} used UV light to illuminate  fluorescent markers on the sensing skin surface and used a time-division multiplexed circuit to switch between UV and white light~\cite{hogan2022finger}. This method not only achieves accurate force perception and slip sensing but also reduces the influence of markers on contact texture detection.

From the perspective of signal processing, the function of the illumination system is mainly to improve the effect of tactile perception and cooperate with the sensing skin and vision system to achieve more functions. For example, RGB lighting with the monocular camera can realize depth reconstruction.

\subsection{Vision system}

With the advancements in optical imaging techniques, cameras with miniaturized sizes are now capable of producing higher-quality images, which opens the door for the development of fingertip visuotactile sensors. According to imaging techniques, vision systems can be categorized into monocular  cameras \cite{yuan2017GelSight}, binocular cameras\cite{cui2021hand}, depth cameras\cite{kuppuswamy2020soft}, and event-based dynamic vision sensors (DVS) ~\cite{naeini2019novel}.

The monocular RGB camera is a widely used imaging method due to its versatility and low cost. Many applications, such as GelSight~\cite{yuan2017GelSight} and Digit~\cite{lambeta2020digit}, utilize monocular RGB cameras for imaging. When selecting a monocular camera, the size, field of view (FOV), focal length, and resolution are crucial factors to consider. The camera's size typically determines the sensor's size, while the FOV and focal length determines the sensor's thickness.

Binocular RGB imaging also is a widely used method for imaging. This technology involves using two cameras to capture images of the sensing skin simultaneously and then calculating depth information through binocular stereo matching. One of the main advantages of this method is that it is not affected by lighting conditions, and only requires intrinsics and extrinsics calibration. However, the short baseline of the binocular camera can make it difficult to achieve high detection accuracy, which is primarily determined by the size of the visuotactile sensor.

In addition to binocular imaging, depth cameras also allow for the detection of depth information on the surface of the sensing skin. However, due to their larger size, they are mostly used in sensors with larger dimensions, such as the Soft-bubble sensors~\cite{kuppuswamy2020soft}. Compared to RGB cameras, depth cameras can provide more stable stereo-depth images and eliminate the need for calibration. But they are more expensive and difficult to promote on a large scale. Additionally, Naeini \textit{et al.} have also explored the use of event cameras as internal sensing components for visuotactile sensors~\cite{naeini2019novel}. These cameras offer low time delay, high dynamics, and sensitivity to slip information. However, they are expensive and have low resolution as well as a poor signal-to-noise ratio.

Similar to sensing skin, the vision system is critical to the signal processing method of the visuotactile sensors. For example, for depth reconstruction, monocular cameras are often combined with photometric stereo methods and binocular imaging is often used when binocular cameras are used. When a depth camera is used, it can be obtained directly from the depth camera.

\section{Signal Processing for Visuotactile Sensor }

Compared to traditional electrical signals~\cite{posada2020innovations}, the visuotactile sensor acquires 2D image signal, allowing for signal processing through image processing algorithms. Signal processing for visuotactile sensors typically involves six key areas: contact area segmentation, 3D reconstruction, force perception, slip detection, mapping and localization, and sim-to-real.

\begin{figure}
	\centering
	\includegraphics[width=0.48\textwidth]{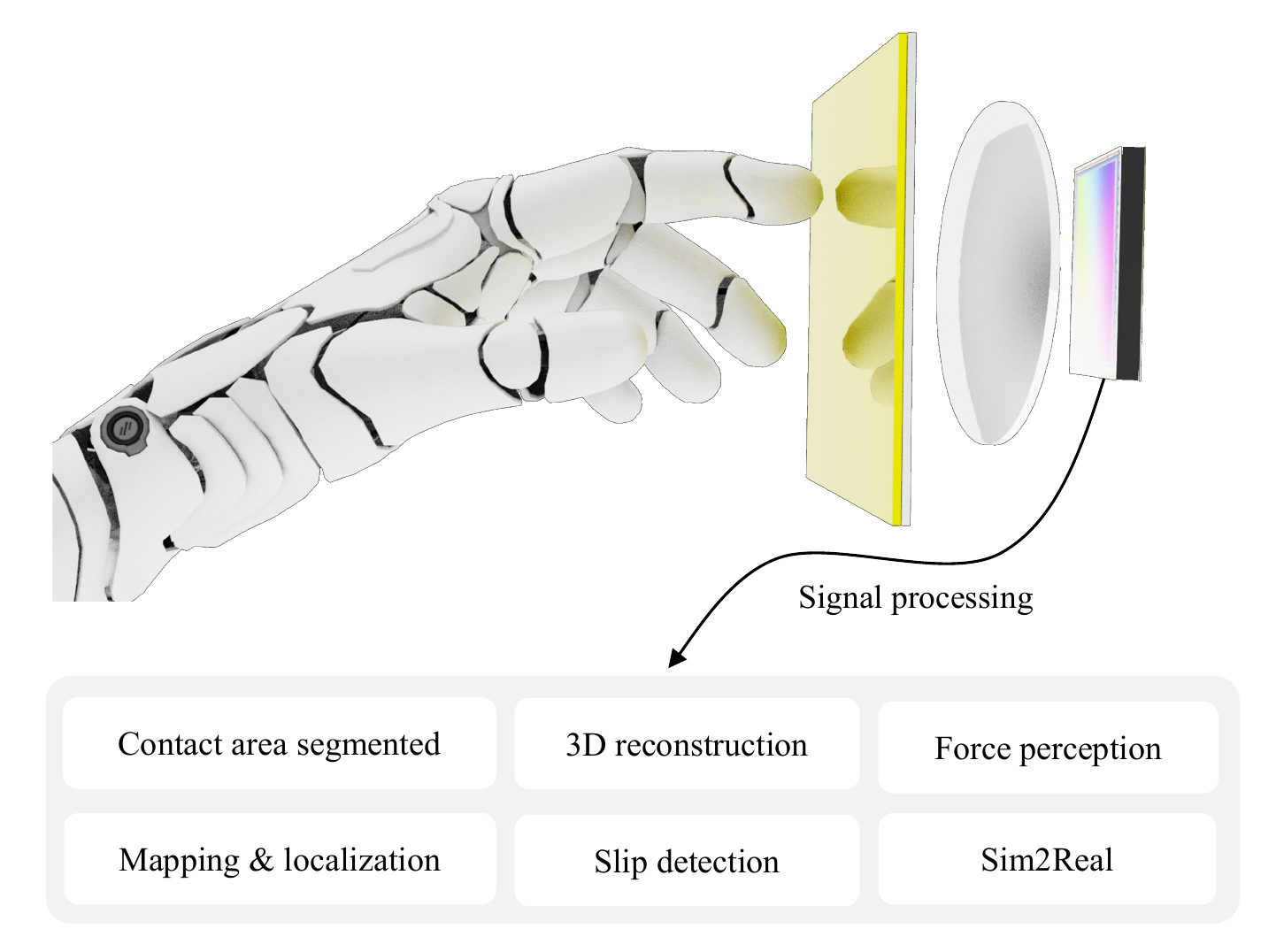}
	\caption{Visuotactile sensor signal processing framework.} \label{fig:4}
\end{figure}

\subsection{Contact area segmentation}

When the visuotactile sensor contacts an object, the sensing skin's color and texture will change. Extracting information about the contact location and area can further improve the stability and success rate of the robot in grasping the object. Generally speaking, there are two primary methods for extracting visuotactile information: traditional image processing methods~\cite{kim2022uvtac} and deep learning methods~\cite{li2021design}. Traditional image processing methods with explicitly mathematical algorithms are usually of fast computational speed, high frame rate, and low latency. Zhang \textit{et al.} used the background difference method for visuotactile information extraction, which involves using image difference to remove the influence of background factors, denoising by erosion and collision, and finally extracting the maximum connected domain as the contact area~\cite{zhang2022tacrot}. However, this method might easily fail at scenes with drastic lighting changes. To address the problem, Li \textit{et al.} proposed a method for extracting visuotactile information in highly dynamic scenes using the TaTa gripper~\cite{li2022tata}. Their approach utilizes a deep learning network with Fully Convolutional Networks (FCN)~\cite{long2015fully} to segment contact regions. While this method offers greater robustness, it requires a significant amount of labeled data and has a slower computational speed.

\subsection{3D Reconstruction}

Compared with contact area segmentation, 3D reconstruction is a more difficult problem. The goal of this task is to generate a dense point cloud of depth information on the sensor surface, which is particularly useful in improving the accuracy of object pose estimation when visual occlusion occurs. This section will discuss mainstream shape reconstruction techniques for visuotactile sensors based on photometric stereo, luminance reconstruction, binocular imaging, structured light \& time of flight (ToF), dense optical flow, and deep learning.

\subsubsection{Photometric stereo methods}

This method is a widely used technique for reconstructing the depth of objects~\cite{santo2017deep,cho2018semi,hernandez2008multiview,goldman2009shape}, which is achieved by mapping the luminance information of pixel points to normal vector information. One key advantage of this method is its low implementation cost, as it can be achieved using RGB cameras. Additionally, it offers high reconstruction accuracy in a small space. However, a major disadvantage is that it requires a high-quality internal light field of the sensor. Therefore, when using this method, it is crucial to design and optimize the sensor's lighting system.

\begin{figure}
	\centering
	\includegraphics[width=0.47\textwidth]{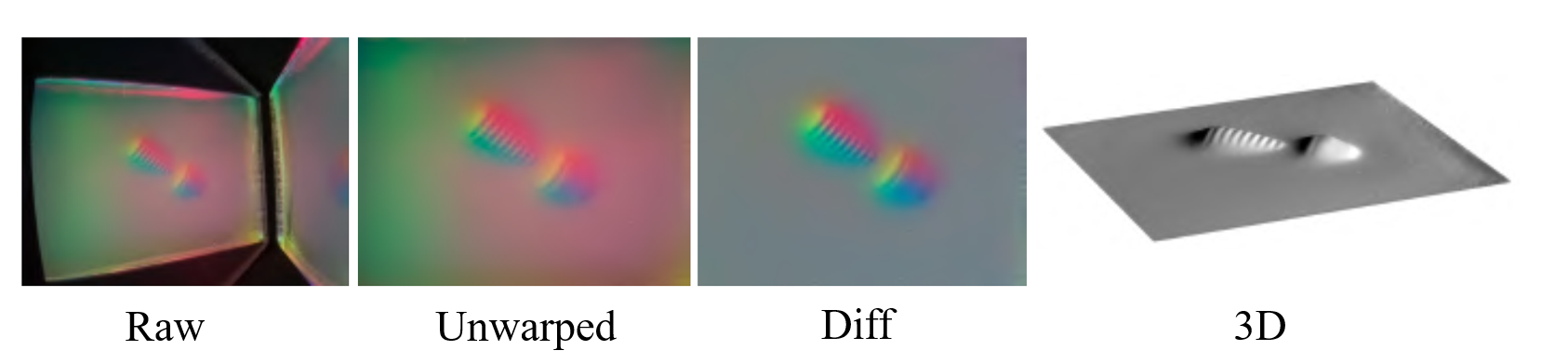}
	\caption{ Reconstruction effect with photometric stereo method~\cite{wang2021GelSight}.} 
 \label{fig:Photometric}
\end{figure}

The photometric stereo method is adapted to the sensing skin that satisfies the condition of Lambert reflection, which has a uniform surface reflection function. It defines the surface of the sensor as $z=f(x,y)$. Assuming that the X, Y coordinate system of the image coincides with the coordinate system of the sensor surface, the gradient $(p,q)$ at the $(x,y)$ point can be expressed as
\begin{equation}
 p=f_{x} =\frac{\partial z}{\partial x},
\label{eqn1}
\end{equation}
\begin{equation}
q=f_{y} =\frac{\partial z}{\partial y} .
\label{eqn2}
\end{equation}
Then the normal vector at the point $(x,y)$ is $(p,q,-1)^{T} $. We assume that there are no cast shadows and reflections on the sensor surface and that its shape and brightness depend only on the normals. Since the sensor surface to be reconstructed is composed of many pixel blocks, and the normal vector of each pixel point indicates its direction, we can complete the reconstruction of the sensor surface information by calculating the normal force of each pixel point. We define the illumination at $(x,y)$ as $I(x,y)=R(p,q)$, where $(p, q)$ is the gradient at $(x, y)$. The reflectance function R(p; q) maps values from a two-dimensional space into a one-dimensional space of intensities. 

The function $R(p,q)$ represents a mapping from a 2D space to a one-dimensional (1D) luminance space. In this case, an intensity value will contain multiple sets of gradient mappings. To eliminate the singularity, we will choose multiple channels for different lighting conditions. In general, we can choose three channels to estimate the pixel distribution, i.e.,

\begin{equation}
\overrightarrow{I} (x,y)=\overrightarrow{R}(p(x,y),q(x,y)),
\label{eqn3}
\end{equation}
\begin{equation}
\overrightarrow{I} (x,y)=(I_{1}(x,y) ,I_{2}(x,y),I_{3}(x,y)), 
\end{equation}
\begin{equation}
\overrightarrow{R} (p,q)=(R_{1}(x,y) ,R_{2}(x,y),R_{3}(x,y)).
\end{equation}

 In this way, we need to establish an expression between the color change and the surface normal~\cite{dong2017improved}. A commonly used calibration method involves using a known-size ball to contact the sensor surface at various locations for sampling. By collecting data on the correspondence between the color change and surface normal, a lookup table can be created, and the optimal solution can be found using search and clustering methods. To enhance the accuracy of the results, Ramamoorth \textit{et al.} utilized a spherical harmonic function for further optimization~\cite{1039204}, which, however, suffered from slow computational speed. To address this issue, Yuan \textit{et al.} employed a fast Poisson solver with discrete sine transform (DST) to accelerate the solving process, enabling parallel computation of the data~\cite{yuan2017GelSight}. Wang \textit{et al.} further optimized the above method by using neural networks instead of the look-up table method and by using the Unet networks~\cite{ronneberger2015u} to achieve depth reconstruction in a two-light case or even with a single light~\cite{wang2021GelSight}, and the results are shown in Fig.~\ref{fig:Photometric}.

\subsubsection{Luminance reconstruction methods}

\begin{figure}
	\centering
	\includegraphics[width=0.47\textwidth]{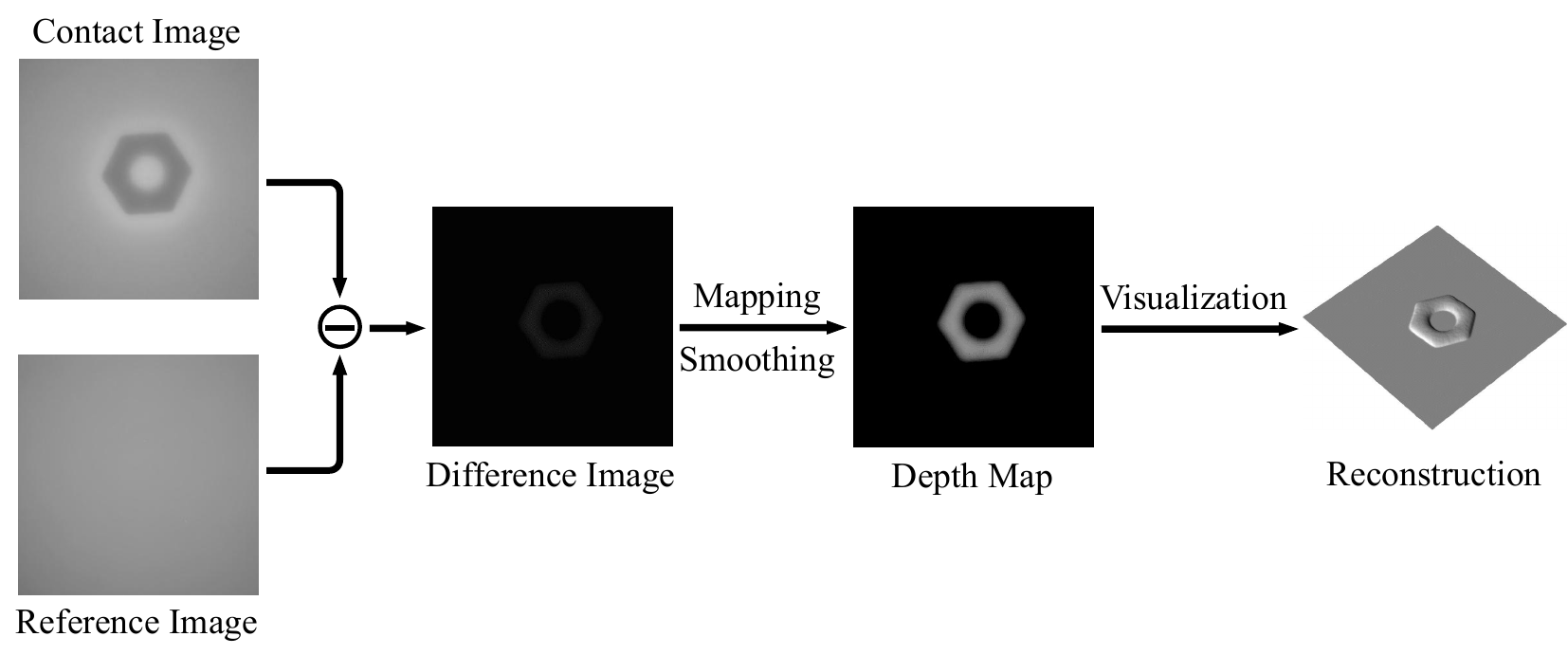}
	\caption{Luminance-based 3D reconstruction method~\cite{lin2022dtact}. This method first uses the background difference method to remove the noise and then calculates the depth based on the luminance information.} \label{fig:6}
\end{figure}

Besides the light field gradient, the luminance can represent depth information as well. As shown in Fig.~\ref{fig:6}, Lin \textit{et al.} designed a translucent membrane that contains a translucent layer and an absorbing layer that not only resists external light but also absorbs light from inside the sensor~\cite{lin2022dtact}. When contact occurs, the deeper the contact area is pressed, the darker the output color will be. Based on this principle, fitting the mapping relationship between the luminance and depth information of each pixel point can realize the reconstruction of the sensor surface information.

\subsubsection{Binocular imaging methods}

Binocular imaging~\cite{grinberg1994geometry,belhumeur1996bayesian,mitiche1985tracking} is also one of the methods to achieve the 3D reconstruction of the sensing skin. Similar to the human eye, binocular imaging can acquire depth information by calculating the disparity between two cameras at different locations when photographing the same object.

\begin{figure}
	\centering
	\includegraphics[width=0.49\textwidth]{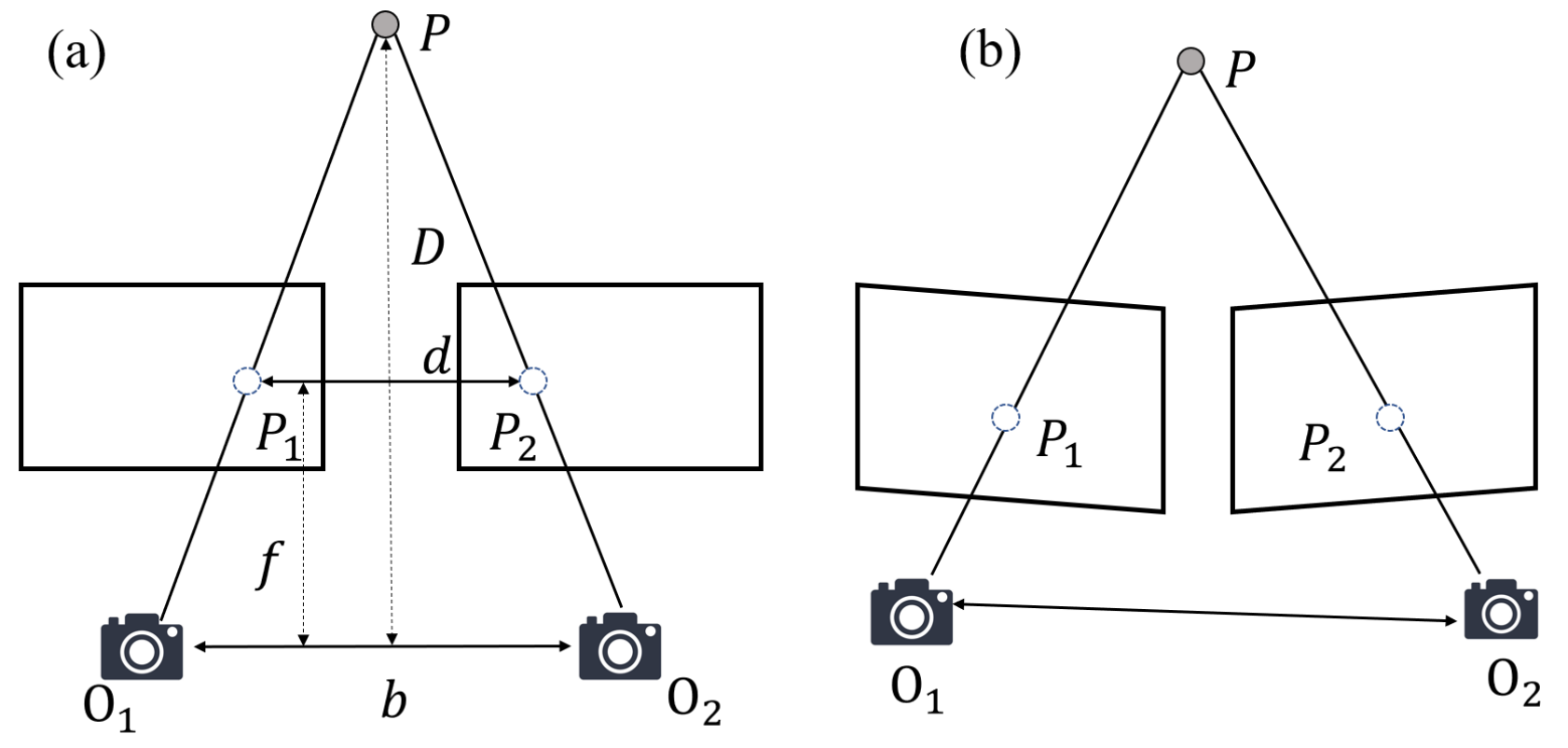}
	\caption{The principle of binocular imaging. (a) Ideal situation: binocular camera imaging planes are parallel to each other. (b) Non-ideal situation: the binocular camera imaging plane has an angular difference.} \label{fig:7}
\end{figure}

The principle of binocular imaging assumes that the baseline of the camera is $b$, the depth of the target is $D$, the focal length is $f$, and the distance difference between the two cameras is $d$, as shown in Fig.~\ref{fig:7}(a). The geometric relationship can be expressed as
\begin{equation}
D=\frac{b\times f }{d}.
\label{eqn4}
\end{equation}

In ideal conditions, the cameras are in a uniform plane, but the optical centers of the two cameras are not in the same plane in most practical cases, as shown in Fig.~\ref{fig:7}(b). To make the depth calculation more convenient, it is necessary to convert the non-ideal conditions to ideal conditions by image correction method~\cite{cui2014precise}. The error of binocular imaging in depth detection $ \bigtriangleup D $ can be expressed as follows
\begin{equation}
D+\bigtriangleup D =  \frac{b\times f}{d+\bigtriangleup d},
\label{eqn5}
\end{equation}
\begin{equation}
\bigtriangleup D= \frac{b\times f}{d} -\frac{b\times f}{d+\bigtriangleup d},
\label{eqn6}
\end{equation}
\begin{equation}
\bigtriangleup D= D- \frac{1}{\frac{1}{D}+\frac{\bigtriangleup d}{b\times f}  }.
\label{eqn7}
\end{equation}

Eqn.~\ref{eqn6} explains that when the baseline and focal length are constant, the accuracy of parallax $d$ determines the accuracy of depth imaging, and smaller parallax deviation brings smaller depth deviation. In addition to parallax, the size of the baseline and focal length will also have an impact on the detection accuracy. Eqn.~\ref{eqn7} shows that a longer baseline and focal length will improve the depth accuracy. Due to the limitation of the size of the optical-tactile sensor, the size of the binocular camera is inevitably small. Therefore, reducing the parallax error become important. Binocular imaging is based on the principle of feature matching, and it is difficult to achieve good detection accuracy for sensor surfaces with few feature points.

To solve this problem, Zhang \textit{et al.} set seven markers on the sensor surface and calculated the change of distance and displacement of each marker by binocular imaging method~\cite{zhang2018robot}. The problem with this method is that the number of markers is too small and it is difficult to accurately reflect the deformation of the sensing skin surface, but the increase in the number of markers will increase the difficulty of marker matching. To achieve dense markers matching, Cui \textit{et al.} proposed a structure-based markers stereo matching method, which first detects markers on the sensor surface and later performs a look-ahead sorting algorithm to match markers in the images captured by the binocular camera~\cite{cui2021hand}. In fact, the monocular can also achieve binocular imaging. Zhang \textit{et al.}  changed the detection direction of the camera through the mirror and acquired images of two mirrors through a single camera. This method not only can get high-precision binocular images but also can reduce the cost~\cite{zhang2022tac3d}. In addition to the matching error, the refraction of the sensing skin also affects the detection. To obtain more accurate depth information, Hu \textit{et al.}  proposed a curved visuotactile sensor GelStereo Palm~\cite{cui2021hand} and used GP-RSRT (Refractive Stereo Ray Tracing model for GelStereo Palm) to solve the refraction problem generated when light passes through the elastomer and air ~\cite{hu2023gelstereo}. After experimental tests, the method can achieve an average perception error of 0.21 mm.

\subsubsection{ToF \& structured light methods}
Compared with binocular imaging, ToF \& structured light methods~\cite{zanuttigh2016time} mostly use the active projection method, so they have higher detection accuracy and interference resistance. ToF is a method that uses the measurement of the light time of flight to obtain distances. This method is highly adaptable and can obtain valid depth information regardless of whether the object has feature points or not, so the method can be applied to the reconstruction of convex surfaces without markers. Structured light ~\cite{hernandez2008multiview,cho2018semi,dhillon2015geometric} is a system that comprises a projector and a camera, used to capture specific light information projected onto an object's surface and background. This information is then analyzed to determine the object's position and depth, thereby reconstructing the entire 3D space. However, ToF \& structured light methods  require high-quality projection equipment and cameras, which can be expensive. 

To reduce costs, researchers often use readily available depth cameras like Intel Realsense~\cite{Realsense} and Pmd~\cite{PMD}. Alspach \textit{et al.}  designed a tactile sensor that can detect up to 15 cm in diameter, using a latex elastic film as the sensing skin and a Pmd CamBoard Pico Flexx camera as the imaging device~\cite{kuppuswamy2020soft}. Li \textit{et al.} improve the Soft-bubble by using the Realsense L515 camera with higher detection accuracy as the sensing device and designed a passively retractable three-finger platform to achieve object grasping. The sensor can achieve object classification and grasp based on tactile information~\cite{li2021design}. The major factor that greatly limits this method to scale up is the high costs. 

 \subsubsection{Dense optical flow methods}

Although binocular imaging can achieve 3D reconstruction, it is difficult to obtain high-resolution depth reconstruction with sparse markers. To solve this problem, Du \textit{et al.} proposed a scheme using a dense color pattern instead of a dot matrix and employed a dense optical flow algorithm to track the deformation of the elastomer surface, which relies on monocular RGB to achieve high resolution and high accuracy depth reconstruction~\cite{du2021high}. Zhang \textit{et al.} further optimized the hardware structure and algorithm and proposed a new generation of the visuotactile sensor, DelTact~\cite{zhang2022deltact}. Li \textit{et al.}  combined binocular imaging with a dense color pattern to design a sensor with a detection accuracy of 10 $\upmu$m and a temporal resolution of 11 ms, which can be used for 3D traction stress measurement~\cite{li2022imaging}.

\subsubsection{Deep learning methods}


Although binocular imaging and the dense optical flow method can achieve good results in 3d reconstruction, they are both only applicable to sensors where makers are present. For the 3D reconstruction of sensors without markers, deep learning is a more general approach that is independent of the sensor surface shape and lighting conditions.

To achieve the depth reconstruction of DenseTact~\cite{do2022densetact} (a 3D visuotactile sensor without markers), Do \textit{et al.}  proposed an adaptive depth information reconstruction network, whose input information is the image captured by the camera and the output is the depth information of the contact location. Nevertheless, this method requires a large amount of reference data (29,200 training data and 1,000 test data are collected in the experiments).

\subsection{Force Perception}

Force perception is one of the most important functions of tactile sensors. Accurate and stable force perception not only improves the manipulation and control of robots but also ensures human-robot interaction safety. For visuotactile sensors, current force perception methods are mainly based on marker detection, finite element modeling (FEM), and deep learning.

\subsubsection{Markers detection methods}

\begin{figure}
	\centering
	\includegraphics[width=0.47\textwidth]{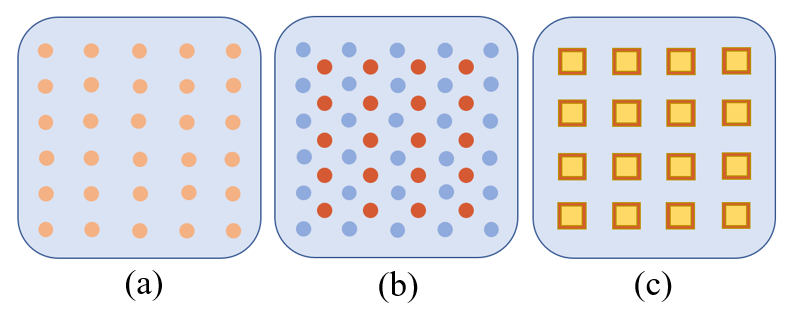}
	\caption{Different types of markers. (a) Single color dot markers array. (b) Dual-color dot markers array. (c) Dual-color square markers array.} \label{fig:8}
\end{figure}

As shown in Fig.~\ref{fig:8}, to improve the detection effect, marker layers of different structures are designed, which will greatly help the sensor sense the forces in different directions.  In practical applications, the tangential and normal forces can be estimated by extracting the change of size and position of the marker, which is called the markers detection method. Obinata \textit{et al.} found that the tangential and normal forces can be obtained by calculating the offset of markers that coupled with each other~\cite{obinata2007vision}. In his experiment, four points in the central region of the sensor were marked in red, the offset of the red makers in the central region was used to represent the tangential force, and the radius of the contact area was used to represent the normal force, which is an intuitive and effective way, but the resolution is low. Afterward, Obinata \textit{et al.} further designed a sensor with a two-layer structure~\cite{sato2008measurement}, where each layer of the sensor has markers of different colors, and the contact force is calculated by detecting the relative offsets of the two markers of different colors. To further improve the spatial resolution of the sensor, Lin \textit{et al.} designed overlapping double-layer square markers based on the principle of diffuse and transmission of light. The shear deformation is determined from the center of mass of the marker, and the normal deformation is obtained by the color change of the markers~\cite{lin2019sensing}.

Apart from hardware, algorithmic optimization can also achieve the decoupling of the two forces. Sato \textit{et al.} proposed a method for normal force, tangential force, and moment decomposition using the Helmholtz-Hodge Decomposition algorithm~\cite{zhang2019effective}, which is commonly used in computational fluid mechanics and can decompose arbitrary optical flow fields into rotational and scattering components. This method has high data efficiency and low complexity, in the real-world experiment, the calibration of the sensor only uses 300 data points. Although the force detection method based on markers has a good detection effect, this method is mainly for visuotactile sensors with markers, and cannot be applied to force sensors without markers.

\subsubsection{FEM methods}
 The relationship between sensor deformation and contact forces can also be studied from the perspective of materials. The FEM methods are designed for tactile sensors with markers, which have a high resolution. Ma \textit{et al.} combined FEM~\cite{bhavikatti2005finite} with markers to predict the deformation of the sensor by using the offset of the markers as input and then estimating the magnitude of the contact force~\cite{ma2019dense}. The method combines information such as Young's Modulus and Poisson's ratio of the sensor surface so that the dense contact force information of the sensor surface can be established with fewer data.

\subsubsection{Deep learning methods}
 To achieve force perception for visuotactile sensors without markers, the deep learning-based force perception approach is employed. Kyung \textit{et al.} proposed a transformer-based contact force detection method for DenseTact~\cite{do2022densetact}, which can segment the contact position and extract force on sensors without markers. However, this method only outputs an overall contact force value for a single image. To a dense contact force heatmap, Sun \textit{et al.} proposed a network~\cite{sun2022soft} that can detect the contact force of each pixel using the ResNet network~\cite{he2016deep}, which can reach a spatial resolution of 0.4 mm with the force detection accuracy of 0.03 N.
 
\subsubsection{Data acquisition}

\begin{figure}
	\centering
	\includegraphics[width=0.43\textwidth]{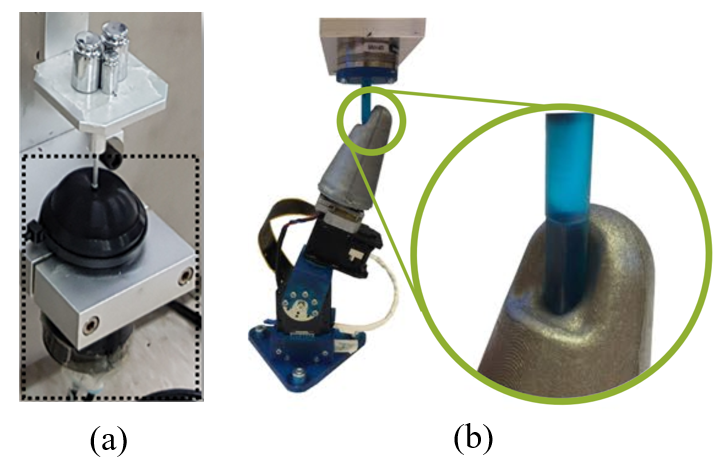}
	\caption{Calibration System. (a) Manual calibration system~\cite{kakani2021vision}. (b) Automatic calibration system~\cite{sun2022soft}.  } \label{fig:calibration}
\end{figure}

Both markers detection methods and deep learning methods are data-driven, so  calibration is an important part of realizing force sensing for visuotactile sensors. Before performing force calibration, researchers need to build a platform containing force sensors, probes, and precision slips. As shown in Fig.~\ref{fig:calibration}, calibration systems can be divided into manual calibration~\cite{kakani2021vision} and automatic calibration~\cite{sun2022soft}. When the amount of collected data is small, the manual calibration system can meet the requirements, but when the amount of data collected is large, the automatic calibration system becomes necessary.

\subsection{Slip detection}
Slip detection technology can improve the stability of the robot in object grasping and operation, especially for manipulating irregularly shaped or fragile objects, where the gripping force and gripping strategy need to be adjusted in time according to the slip signal\cite{romeo2021method}. Moreover, slip detection can be applied to human-computer interaction and virtual reality scenarios. This is because slip signals contain dynamic and detailed interaction information, and the efficiency of human-computer interaction can be improved by slipping commands. Slip information can be obtained from different physical quantities, such as vibration~\cite{fernandez2014micro}, temperature~\cite{accoto2012slip}, tangential force~\cite{melchiorri2000slip}, etc. The most common method for visuotactile sensors is to obtain slip information based on the displacement of the surface marker.

\begin{figure}
	\centering
	\includegraphics[width=0.47\textwidth]{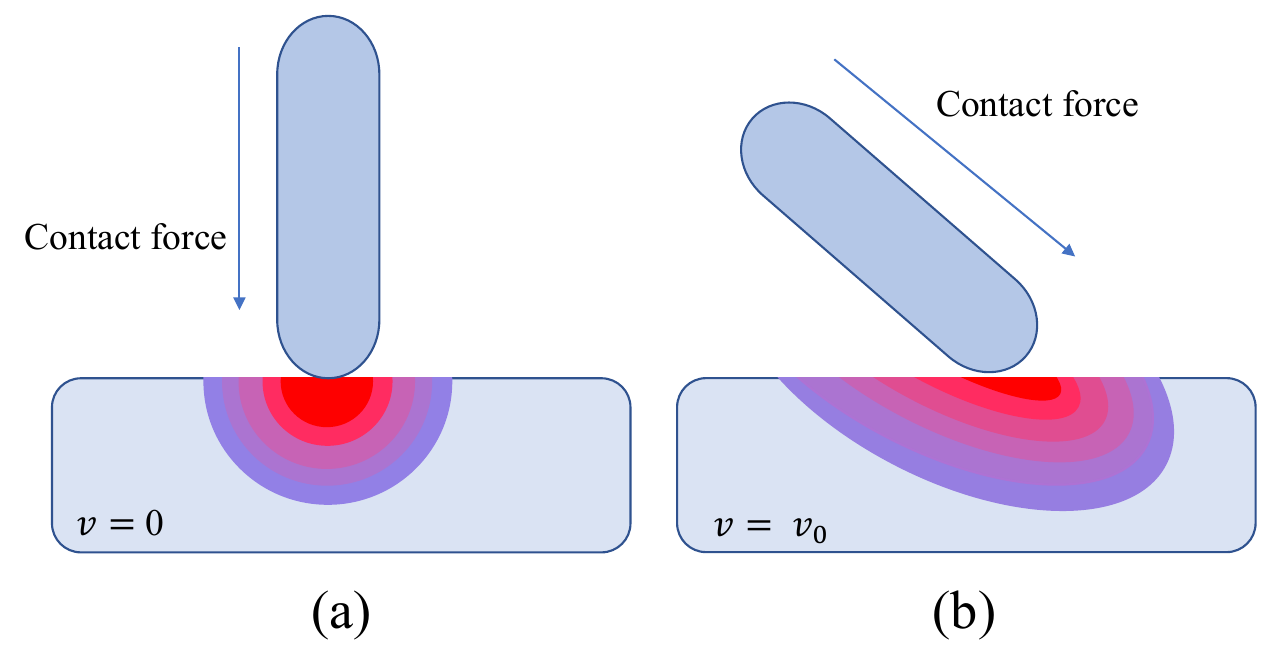}
	\caption{The process of slip occurs. (a) When the force is perpendicular to the contact plane, no slip occurs. (b) Slip will occur when there is a horizontal component of the contact force, and the horizontal force is greater than the frictional force. From red to blue indicates the contact force from large to small.}\label{fig:slip}
\end{figure}

However, the displacement of the marker not only combines the slip information but also reflects the tangential and normal forces.  
As shown in Fig.~\ref{fig:slip}, slip occurs when the tangential force on the sensor surface is greater than the frictional force.
To extract the slip information, Watanabe \textit{et al.} proposed the slip margin measure of "stick ratio"~\cite{watanabe2008grip}, which compares the difference between the displacement of the sensor center point and the displacement of the stick region. To verify the effect of slip information on improving the grasping success rate, a slip detection experiment was designed and the experimental results surface that the slip signal is very helpful in improving the grasping success rate. Yuan \textit{et al.} further analyzed the relationship between the displacement of markers and shear, partial slip, and slip, and proposed a method to determine whether slip occurs based on the entropy of the displacement field offset distribution~\cite{yuan2015measurement}. They found that the more inhomogeneous the distribution of the displacement field, the higher the entropy, and the higher the possibility of slip, but this method is only effective when the surface of the contact object is flat and the texture of the contact surface is small.

Dong \textit {et al.} proposed a method to detect slippage by tracking the relative displacement of the markers and the object~\cite{dong2017improved}. Slip is considered to have occurred when there is a significant displacement of the contact position between the markers and the object. The method was tested on 37 objects and achieved a slip detection accuracy of 71\%. Dong \textit{et al.} further analyzed the causes of slip occurrence from the physical and mechanical perspectives. Under the assumption that the object in contact is a rigid body and the motion of the object on the sensor surface is a 2D rigid body motion, the deviation of the real motion field detected by the sensor from the rigid change of the 2D plane is used as the basis for judgment~\cite{dong2019maintaining}. This method can achieve slip detection without any prior knowledge, 240 tests have been performed on 10 objects, and the detection accuracy can reach 86.25\%. Sui \textit{et al.}  proposed an incipient slip detection method based on the force and deformation distribution information of the sensor, which initially determines the central region of the rod by the force distribution, and later detects the direction and magnitude of slip in the whole contact region. To verify the effectiveness of the method, they compared the actual scene with the finite element analysis software, and the relative error of detection was within 10\%~\cite{sui2021incipient}. The slip detection method combined with finite element analysis has better interpretability and stability, providing reliable theoretical support for the understanding of the mechanism of slip generation.

Data-driven slip detection is a research hot spot, which has better generality but requires a large amount of data. Zhang \textit{et al.} proposed a slip detection network based on LSTM~\cite{zhang2018fingervision}, the model takes a sequence of 10 groups of sensors as input, and each sequence contains a deformation field and its projection on the x and y axes. In the experiment, 12 daily objects were tested and the classification accuracy reached 97.62\%. James \textit{et al.} proposed a support vector machine-based slip detection method and applied the algorithm to the Tactile Model O (T-MO) robotic hand. To test the effectiveness of the slip detection algorithm in a real-world scenario, two experiments are designed: one is to make the object slip by adding heavy content to the grasped container, meanwhile using the slip sensing algorithm to detect the occurrence of the slip and adjust the grasping force in time. The other is to test the minimum force required to grasp an object by slip detection. Visual-tactile fusion provides a new solution to the problem of detecting object slipping.  Li \textit{et al.} proposed a slip detection method based on deep neural networks (DNN), which takes 16 images of visual and tactile sensations as an input sequence~\cite{li2018slip}. To verify the detection effectiveness of the algorithm, more than 120 grasping experiments were performed and achieved a detection accuracy of 88.03\%.

In addition to the algorithm design, hardware optimization can also improve the effectiveness of slip detection. Maldonado \textit{et al.}  designed a finger with a hole to detect the texture and distance of an object by placing a micro sensor inside the hole~\cite{maldonado2012improving}. When the object slips along the contact area, the texture of the object detected by the sensor will change. According to this principle, we can determine whether a slip has occurred. However, the disadvantage of this design is that it cannot detect slips in contact with smooth or transparent objects.

\subsection{Mapping and localization}
\begin{figure}
	\centering
	\includegraphics[width=0.47\textwidth]{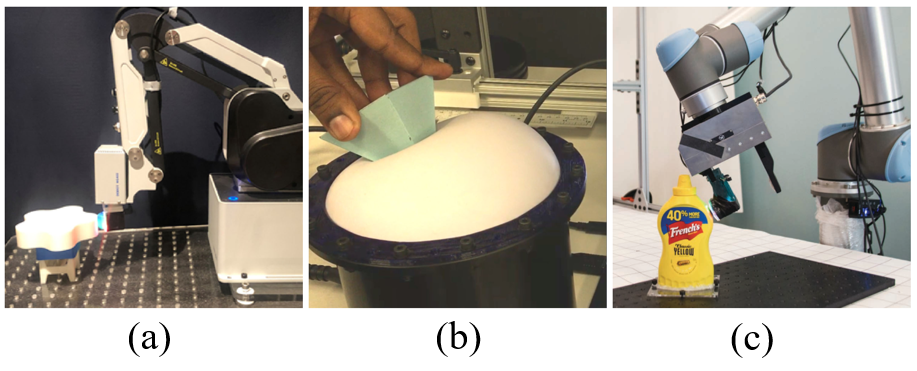}
	\caption{Mapping and localization. (a) Edge contour detection~\cite{lepora2022digitac}. (b) In-hand object pose estimation~\cite{kuppuswamy2019fast}. (c) Object pose estimation in the scene~\cite{wang20183d}.  } \label{fig:Mapping}
\end{figure}

Mapping and localization is a crucial technique used to determine the position and orientation of an object in a coordinate system. This method finds extensive applications in various fields such as grasping, navigation, and augmented reality. In addition to reconstructing the 3D information of the contact area, visuotactile sensors can also leverage priori knowledge to calculate the position and pose of the object being touched. As shown in Fig.~\ref{fig:Mapping}, mapping and localization of objects can be broadly classified into three categories: edge contour detection, in-hand object pose estimation, and object pose estimation in the scene.

\subsubsection{Edge contour exploration}

Exploring object contours through tactile perception is meaningful for enabling object grasping in low-visibility scenarios. However, small-size visuotactile sensors such as GelSight and Digit are limited in their ability to acquire global information about objects via a single contact.

Lepora \textit{et al.} proposed a deep learning approach for achieving object contour exploration~\cite{lepora2019pixels}, which involves using neural networks to extract the contour of contact between the TacTip~\cite{ward2018tactip} and the object, and by edge following to achieve object shape perception. This approach is effective in accurately perceiving the shape of objects through tactile sensing. A similar work was presented on surface following using a GelSight sensor in \cite{lu2019surface}. In a recent study by Lepora \textit{et al.}, an optimized version of the previous model was proposed, called PoseNet~\cite{lepora2022digitac}. This deep learning-based tactile servo control model is capable of detecting the contours of surfaces and edges of objects. The authors tested the model's generality by applying it to three different sensors, namely Digit~\cite{lambeta2020digit}, DigiTac~\cite{lepora2022digitac}, and TacTip~\cite{ward2018tactip}.

\subsubsection{In-hand object pose estimation}

Visuotactile sensors can also improve the accuracy of object pose estimation. 
The estimation of the pose of an object in hand is one of the challenging topics in the field of robotics. Since the fingers of the robot will block the object when the gripper grasps it, it is difficult to estimate the object's pose accurately by vision. However, the application of visuotactile sensors further promotes the development of in-hand pose estimation of objects. Bauza \textit{et al.} proposed a tactile sensing method for in-hand object localization, first establishing a mapping of tactile and object local shapes through a data-driven approach, followed by object localization through a CTI-ICP-N approach, which combines closest tactile imprint (CTI) with ICP iterative closest point (ICP)~\cite{bauza2019tactile}. Here, N denotes the number of closest images matched for the first time based on tactile information. However, this approach requires collecting a large amount of data. To reduce the workload of data collection, Villalonga \textit{et al.} proposed a method to establish a mapping between tactile impressions and local shapes from the simulator and used data augmentation to reduce the differences between real scenes and simulated data~\cite{villalonga2021tactile}. To detect object in-hand pose changes in the presence of occlusion, Anzai \textit{et al.} proposed a deep gated multi-modal learning method, which can be generalized to unknown objects~\cite{anzai2020deep}. Kuppuswamy \textit{et al.}  proposed a step-wise in-hand object pose estimation method based on Soft-bubble~\cite{kuppuswamy2020soft}, which first uses the forward model to predict the deformation of the object when contact occurs with the gripper, and then uses the inverse model to extract the region where the contact between the sensor and the object occurs, and finally uses ICP to achieve the pose estimation of the in-hand object~\cite{kuppuswamy2019fast}. Prior works using tactile array sensors to estimate the object pose \cite{bimbo2016hand} or localize the contact \cite{luo2015localizing} could also be applied with visuo-tactile sensors.

\subsubsection{object pose estimation in the scene}

For object pose estimation in the scene, vision detection is the mainstream method. Although vision detection has good results in obtaining the outline information of the object, it is very difficult to detect some detailed texture information. And the addition of tactile information will be a major help in improving detection accuracy.

Wang \textit{et al.}  proposed a tactile-assisted object monocular depth reconstruction method, which initially roughly reconstructs the outline of the object by monocular vision, and then updates and optimizes the outline information of the object by tactile feedback~\cite{wang20183d}.
Suresh \textit{et al.} proposed a Monte Carlo-based global localization method for contact position, which can obtain the position and information of the sensor relative to the object based on the position of the sensor in contact with the object, and record the movement path of the sensor ~\cite{suresh2022midastouch}. Chaudhury \textit{et al.} built a perception platform with a depth camera, color camera, and tactile sensors, and improved the accuracy of object pose estimation by collocated image~\cite{chaudhury2022using}. The detection method first finds the target object guided by visual detection, afterward uses the depth image to estimate the object's pose, and finally, the pose is calibrated using the tactile sensor.

\subsection{Sim-to-real}

Reinforcement learning ~\cite{sikander2021reinforcement,han2023survey,elguea2023review} offers innovative approaches for tackling complex robot control tasks in challenging environments. However, the low sampling efficiency of the learning approach can jeopardize the equipment's durability in real-world scenarios. And model training demands high-quality large datasets to ensure reliability. To address these issues, the imaging principle of the sensor has been leveraged to simulate the signal generation process using a simulation engine. This approach enables the collection of a significant amount of useful data in a short time and overcomes the problem of sensor aging.
Gomes \textit{et al.}  proposed a tactile information simulation method in Gazebo~\cite{koenig2004design} that simulates the optics in a real scene using the Phong's shading model~\cite{gomes2021generation}. The method first captures the depth map of the object surface through the depth camera in the simulator and then acquires the height map of the deformed membrane by applying bi-variate (2-D) Gaussian filtering.

However, this method mainly considers the projection of light and does not take into account the physical properties of refraction and reflection of light in the process of propagation. To get more realistic contact information,  Agarwal \textit{et al.} proposed a rendering optical simulation system based on the physics-based rendering (PBR), which allows more flexibility in modifying the optical properties of lights, cameras, and elastic films~\cite{agarwal2021simulation}. This approach allows for more realistic simulation images but requires high-performance computers. To improve the speed of calculation, Wang \textit{et al.} used PyBullet~\cite{coumans2016pybullet} as the physical interaction software to perform light rendering and post-processing of contact information through OpenGL~\cite{shreiner2009opengl}, which is fast, flexible, powerful, and supports rendering shadows to obtain more realistic simulation data~\cite{wang2022tacto}. Most previous studies have focused on how to implement the transition from simulation results to the real world (Sim-to-Real), Jianu \textit{et al.} bridged the simulation-reality gap by learning the surface artifacts from real data via a CycleGAN network~\cite{zhu2017unpaired}, which was extended in \cite{jing2023unsupervised}. Chen \textit{et al.} designed a bi-directional generator that can implement Real-to-Sim and Sim-to-Real~\cite{chen2022bidirectional}, which also uses the Domain Adaptation method based on CycleGAN~\cite{zhu2017unpaired}.

Although the above methods can obtain more realistic simulation data, they are achieved purely by optical rendering and do not take into account the physical deformation of the sensor in contact with the object. Chen \textit{et al.} built a visuotactile sensor simulation environment using the Taichi~\cite{zhu2017unpaired}, an open-source computer graphics language that can be used in 3D object simulation and physics simulation, which is not only compatible with Python but also has high computational efficiency~\cite{chen2023tacchi}.

Apart from simulating the deformation information during contact, force information can also be obtained from the simulation. Si \textit{et al.} proposed a framework that combines the marker's motion field of sensor elastic deformation with optics, which accurately simulates the texture information during contact and achieves the simulation of the marker's motion field~\cite{si2022taxim}. Xu \textit{et al.} proposed a penalty-based tactile model to calculate the mechanical information generated by the contact between each point and the object in the simulation environment, the method can not only generate tangential and normal forces but also achieve a computational speed of 1,000 frame/s~\cite{xuefficient}. To evaluate the simulator performance, they implemented a peg-insertion task by the method of data migration, and achieve an 83\% success rate in the real world when trained entirely based on the simulation environment.

To further improve the versatility of the simulation system, Church \textit{et al.} developed a Sim-to-Real and Real-to-Sim deep learning framework based on the gym~\cite{brockman2016openai} simulation environment~\cite{church2022tactile}. Still, initially, this framework was developed for TacTip~\cite{ward2018tactip}. Then, to improve the generality of the framework, Lin \textit{et al.} developed Tactile Gym 2.0~\cite{lin2022tactile}, which can be adapted to TacTip~\cite{ward2018tactip}, Digit~\cite{lambeta2020digit}, and DigiTac~\cite{lepora2022digitac}. Recently, Gomes~\cite{gomes2023beyond} investigated how to simulate light paths in curved surfaces, with validation of simulating the highly curved GelTip sensor \cite{gomes2020geltip}.  

\section{Application of Visuotactile Sensors}

In this section, we will introduce the applications based on visuotactile sensors. With a large sensing area and high resolution, the visuotactile sensors can achieve many challenging tasks such as fabric classification, shape classification, peg-in-hole insertion, etc.

\subsection{Classification}

\begin{figure*}
	\centering
	\includegraphics[width=1\textwidth]{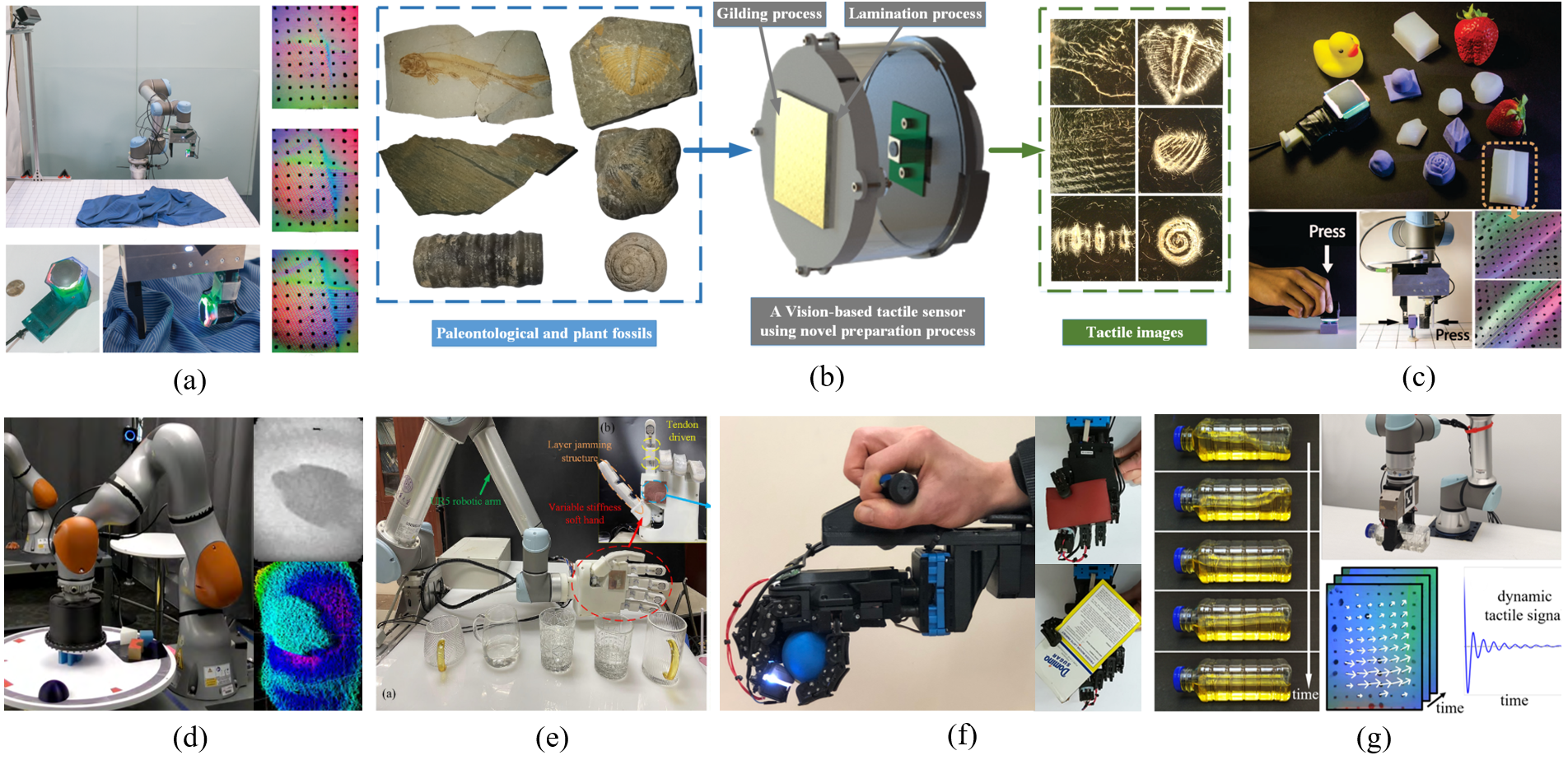}
	\caption{Application of visuotactile sensors in classification. (a)  Clothing material classification~\cite{yuan2018active}. (b) Fossil classification~\cite{zhang2023novel}. (c) Hardness classification~\cite{yuan2017shape}. (d) Shape classification~\cite{alspach2019soft}. (e) Classification of transparent objects~\cite{zhang2022multimode}. (f) Classification of objects in hand~\cite{ward2020miniaturised}. 
 (g) Liquid classification~\cite{huang2022understanding}. } \label{fig:classification}
\end{figure*}

Fabrics are common items in daily life, but their classification is very challenging because they not only have different textures but also different patterns. To obtain more detailed texture information, Yuan \textit{et al.} used visuotactile perception technology and visual perception technology to improve the classification accuracy of fabrics~\cite{yuan2017connecting}. They collected a large amount of training data using GelSight and RGB cameras. And using visual, tactile, and visual-tactile fusion methods for fabric classification, they demonstrated that the addition of tactile perception effectively improves the classification accuracy of fabrics. In \cite{luo2018vitac}, the correction of features in visual and tactile data of fabric textures was maximized so as to weakly pair visual and tactile perception. However, in both works the process of tactile data and visual data acquisition process is very tedious. To achieve automated data acquisition and classification, as shown in Fig.~\ref{fig:classification}(a), Yuan \textit{et al.} improved the previous method by actively perceiving the type of clothes by touch, which first obtains the appropriate grasping position by vision, and then using a visuotactile sensor to obtain texture information~\cite{yuan2018active}. The method can acquire 11 attributes of test cloth samples such as clothing thickness, hardness, fuzziness, etc., and achieves 73\% classification accuracy on 153 fabrics. Some recent works also investigated spatio-temporal attention \cite{cao2020spatio} or cross-modal perception \cite{lee2019touching} in fabric texture perception.  In addition to fabric classification,  In addition to fabric classification, Fang \textit{et al.}  proposed a fabric defect detection method based on visuotactile sensors, which can achieve close to 100\% detection accuracy.
Similar to cloth, the classification and detection of fossils are equally challenging. Fossils gradually lose their texture in weathering, so it is difficult to achieve accurate texture detection relying on visual inspection. As shown in Fig.~\ref{fig:classification}(b), to improve the classification accuracy of fossils,  Zhang \textit{et al.} optimized the elastic film by metal foil plating process and achieved 100\% classification accuracy in the experimental test~\cite{zhang2023novel}. 

Visuotactile sensors can also be used for hardness classification. Yuan \textit{et al.} overcomes the influence of object shape and texture on hardness classification~\cite{yuan2017shape}. They designed a recursive neural network that uses the video sequence of GelSight and object contact (Fig.~\ref{fig:classification}(c)) as input. This method can achieve hardness recognition of objects with similar shapes, but there are limitations for some objects with complex shapes or spine surfaces. In addition to hardness classification, Chen \textit{et al.} applied visuotactile sensors to the field of fruit ripeness classification, which was used to determine the ripeness and health status of fruits by obtaining their hardness and surface characteristics, and achieved a classification success rate of over 92\%~\cite{chen2022non}.

Object shape perception is also a characteristic application of the visuotactile sensors. Limited by the size of the sensor, it is unlikely to obtain all the information about the contacted object at one time. To solve this problem, contour tracking algorithms are proposed. As shown in  Fig.~\ref{fig:classification}(d), Alspach \textit{et al.} designed a visuotactile sensor with a perceptual diameter of 150 mm which utilizes a latex film as the elastic surface~\cite{alspach2019soft}. The sensor expands the elastic membrane by inflating it to obtain greater sensing depth. This large-area, high-resolution visuotactile sensor can acquire the texture, and shape of an object through a single touch. In addition, visuotactile sensors can also be integrated into a multi-finger gripper. Zhang \textit{et al.} designed a five-finger gripper with a visuotactile sensor as palm (Fig.~\ref{fig:classification}(e)), which has the capability of both texture and temperature detection. Based on this gripper, they proposed a multimodal fusion method for transparent object classification, which can achieve close to 100\% classification accuracy for attributes such as style, transparency, and temperature of transparent objects, and 98.75\% accuracy for texture recognition.
Ward \textit{et al.} mounted visuotactile sensors on the fingertips of a five-fingered gripper as Fig.~\ref{fig:classification}(f) depicted and achieved object classification by acquiring tactile information when grasping objects~\cite{ward2020miniaturised}. 

Visuotactile sensors also enabled many creative classification applications. Huang \textit{et al.} proposed a liquids viscosity and volume prediction scheme~\cite{huang2022understanding}, as shown in Fig.~\ref{fig:classification}(g). To achieve liquid property prediction, they introduced a physical model to analyze the oscillation signal and estimate the liquid properties by a Gaussian Process Regression (GPR) model. This method can achieve a classification accuracy of 100\% for water, oil, detergent, etc. The height regression accuracy of sugar water can reach 0.56 mm and the concentration regression error is 15.3 wt\%. In addition, Hanson \textit{et al.} designed a parallel gripper with a spectrometer that enables the classification of liquids by analyzing spectra~\cite{hanson2022slurp}.

\subsection{Grasping}

\begin{figure*}
	\centering
	\includegraphics[width=1\textwidth]{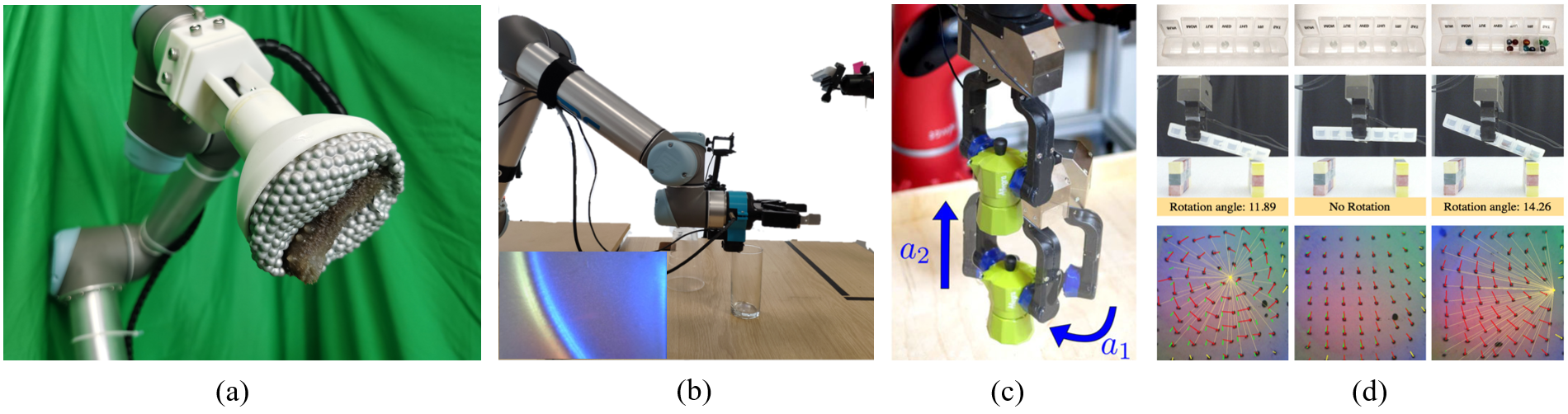}
	\caption{Application of visuotactile sensors in grasping. (a) Underwater object grasping~\cite{li2022tata}. (b) Transparent object grasping~\cite{jiang2022shall}. (c) Tactile perception-based grasping strategy~\cite{calandra2018more}. (d)  Gravity distribution perception~\cite{kolamuri2021improving}.} \label{fig:grasp}
\end{figure*}

Grasping~\cite{saxena2008robotic,miller2004graspit,kehoe2013cloud} is a basic and important function of robots, which can be widely used in garbage sorting, assembly line handling, and home service. Most of the current grasping tasks are done by vision, but for some low-visibility environments with low light or smoke, it is difficult to achieve object detection relying only on visual perception, and the development of visuotactile perception technology has given a great impetus to improve the application range of robots. To solve the problem of object grasping under low visibility, Li \textit{et al.} proposed a gripper with a large detection area and high resolution of tactile perception capability named TaTa~\cite{li2022tata}, which utilizes the refractive index matching principle and particle blocking grasping principle to achieve universal object grasping, as shown in Fig.~\ref{fig:grasp}(a).

Besides low-visibility scenes, the detection of transparent objects is also a major difficulty in the field of vision detection. Transparent objects have special optical properties that not only have less texture information but also lose their depth information in depth cameras. To solve the transparent object grasping problem, Jiang \textit{et al.} proposed a vision-guided transparent object grasping framework, which firstly obtains the poking point by segmenting the network, and then uses GelSight to obtain the tactile information of the point and generates the grasping action~\cite{jiang2022shall}, as Fig.~\ref{fig:grasp}(b) shows. However, this method can only be applied to objects with prior information. To achieve the grasping of unknown objects, Li \textit{et al.} proposed a visual-tactile fusion transparent object grasping and classification framework, which first detects the general position of the object by vision, then calibrates the grasping position using touch and finally achieves the classification of transparent objects using vision-touch fusion~\cite{li2022visual}. After experiments, the framework improves the success rate of grasping transparent objects in complex backgrounds by 36\% and the classification rate by 39\%. Besides visual-tactile fusion, Li \textit{et al.} combined vision, touch, and hearing to help robots achieve object grasping in more complex situations such as stacking, which proved the importance of multimodal perception to solve robot grasping in chaotic scenarios~\cite{li2022see}.

Visuotactile perception not only achieves the classification of texture, hardness, and shape but also perceives force and slip information. To improve the grasping success rate, Calandra \textit{et al.} proposed an end-to-end action state model based on visuotactile perception, which evaluates the current grasping state and the next candidate action to decide the next step to be taken~\cite{calandra2018more}, as shown in Fig.~\ref{fig:grasp}(c). This method improves the robot's grasping ability in three main ways: 1. Increase the grasping success rate. 2. Reduce the number of grasping position adjustments. 3. Achieve object grasping with minimal force. Besides the grasping position, Kolamuri \textit{et al.} considered the effect of object mass distribution on the grasping success rate. As shown in Fig.~\ref{fig:grasp}(d), they proposed a closed-loop grasping system that prevents imbalance when grasping objects with uneven gravity~\cite{kolamuri2021improving}. The system estimates the gravity distribution of the object and adjusts the grasping position using visuotactile perception.

\subsection{Manipulation}

Our human hand with precise force control and dexterity helps accomplish many daily life tasks. The application of visuotactile sensors allows robots better adapted to complex manipulation tasks safely and reduce decision errors.

\begin{figure*}
	\centering
	\includegraphics[width=1\textwidth]{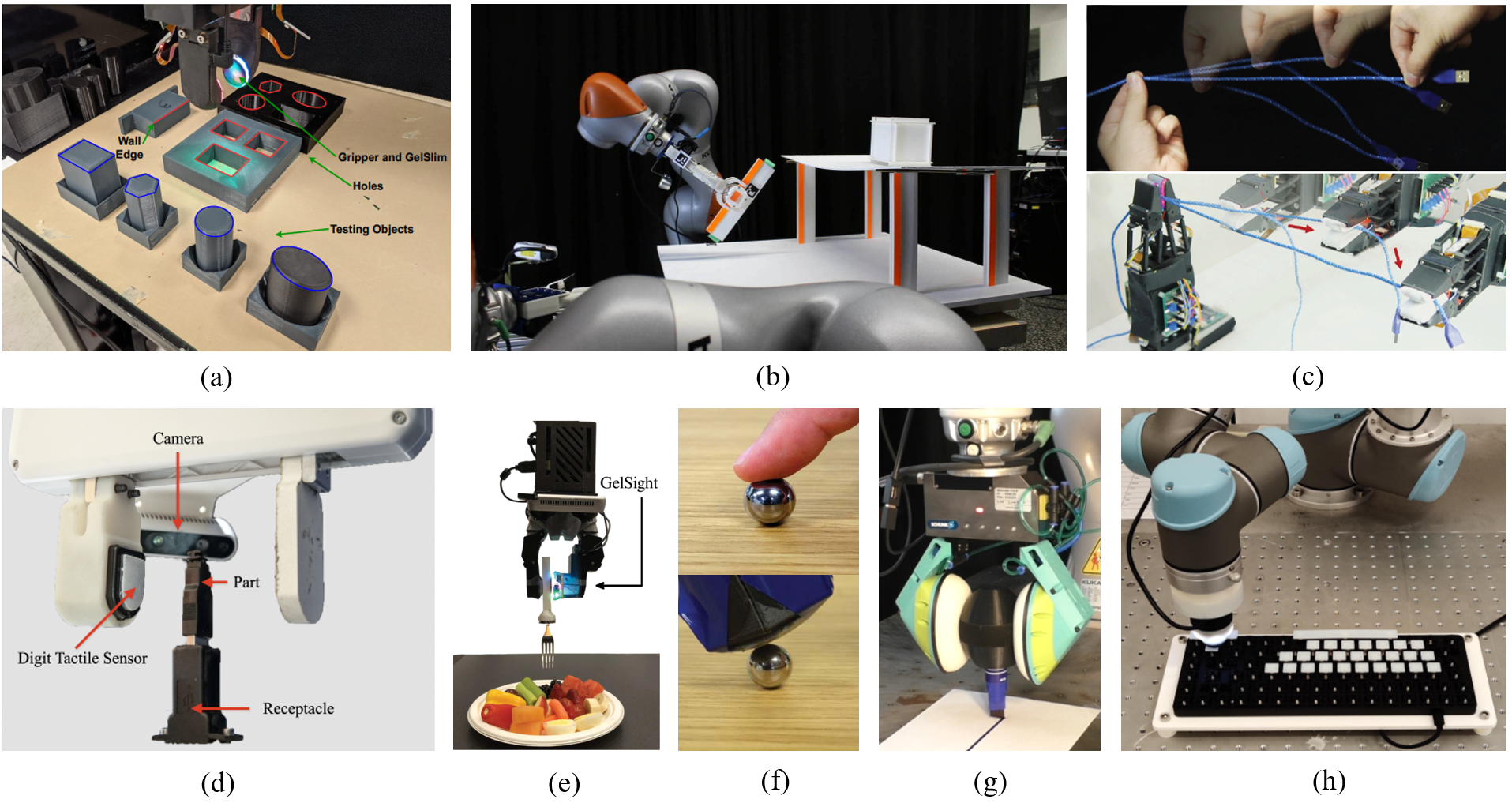}
	\caption{Application of visuotactile sensors in manipulation. (a) Peg-in-hole insertion~\cite{kim2022active}.   (b)  Construction assembly~\cite{belousov2019building}.
    (c) Cable manipulation~\cite{she2021cable}.  (d) USB plug manipulations~\cite{fu2022safely}.
 (e) Fruit manipulation~\cite{song2019sensing}.   (f)  Fingertip manipulation~\cite{tian2019manipulation}.  
 (g) Tool manipulation~\cite{oller2022manipulation}.(h) Keyboard input~\cite{church2020deep}.     } \label{fig:manipulation}
\end{figure*}

Peg-in-hole insertion is a scenario task in workpiece assembly, which is difficult for novice operators. To solve this problem, Kim \textit{et al.} proposed a two-step operation strategy using a gripper with GelSlim~\cite{taylor2022gelslim} as actuator~\cite{kim2022active}, as shown in Fig.~\ref{fig:manipulation}(a). The strategy first uses a tactile model to estimate the contact line between the object and the insertion hole, and later uses a reinforcement learning model to adjust the pose of the object. The experiment shows the method has more than 95\% insertion success rate. As shown in Fig.~\ref{fig:manipulation}(b), visuotactile perception technology can also be applied in the construction field, Belousov \textit{et al.} designed a controller based on marker deviation and proximity vision using FingerVision~\cite{zhang2018fingervision} and applied the controller to construction assembly~\cite{belousov2019building}. Combined with FingerVision's multimodal perception capabilities, it enables tasks such as force following, rotation, and handover, demonstrating a wide range of application scenarios for robots in the construction industry. 

Cable manipulation is one of the hot issues in industrial research. Due to the soft material of cables, it is difficult to build accurate models \cite{pecyna2022visual}. To solve this problem, She \textit{et al.}  using GelSight~\cite{she2021cable} designed a cable manipulation framework based on LQR control and PD control. As shown in Fig.~\ref{fig:manipulation}(c), compared with the open-loop operation, this approach has a faster speed and a higher success rate. To verify the feasibility of the method, She \textit{et al.} also conducted experiments on the operation of many different types of cables with good results, showing that the visuotactile sensors have a wide range of applications in the field of flexible object manipulation. 

Similar to cable manipulation, Sunil \textit{et al.}  developed a clothing manipulation framework using visual-tactile sensors, which first uses vision to obtain the grasping position, and later uses touch to identify and adjust the grasping position~\cite{sunil2023visuotactile}. The framework can achieve the task of folding and hanging clothes by grasping and sliding. As shown in Fig.~\ref{fig:manipulation}(d), Li \textit{et al.} proposed a tactile-based assembly technique that employs a tactile feature-matching algorithm to achieve fine-grained manipulation of fine components, e.g., USB connector insertion~\cite{li2014localization}. This method is simple and feasible, but less generalizable. To address the generalization problem, Fu \textit{et al.} proposed a safety learning strategy with tactile feedback to achieve accurate insertion under the premise of avoiding collision between the robot and the environment~\cite{fu2022safely}. After experiments, the method achieved insertion in 45 different USB plug poses.

Visuotactile sensors can also be used in caring and elderly assistance applications, Song \textit{et al.} applied a visuotactile sensor to food manipulation, using the sensor to obtain the contact force during gripping~\cite{song2019sensing}. As shown in Fig.~\ref{fig:manipulation}(e), the difficulty of this operation is that different foods have different hardness and weight, and it is important not only to ensure that the fork is inserted into the object during the operation but also to detect whether the operation is successful. To solve this problem, Song \textit{et al.} developed a control strategy that utilized the non-linear strain-stress relation of the elastomer to equalize the relationship between the force range and sensitivity.

To demonstrate the advantages of visuotactile perception, Tian \textit{et al.} proposed a tactile Model Predictive Control(MPC)-based control framework to simulate the operation of human fingers when turning a steel ball (Fig.~\ref{fig:manipulation}(f)) or a sieve, which can achieve object position adjustment in the presence of visual occlusion by rolling the object~\cite{tian2019manipulation}. Furthermore, Suh \textit{et al.} applied visuotactile sensors to the squeegee, scribing operation~\cite{suh2022seed}.  To achieve precise control, a force-position hybrid controller was designed, which uses a soft-bubble large sensing surface to acquire the tool's pose and tactile feedback to adjust the contact force between the tool and the environment. This strategy has higher stability compared to open-loop operations. As shown in Fig.~\ref{fig:manipulation}(g), Oller \textit{et al.} modeled the Soft-Bubbles film using a kinetic model and predicted the pose of the manipulated object by the deformation of the film. This method can manipulate many different objects such as pens, spatulas, and sticks~\cite{oller2022manipulation}.

In addition, visuotactile sensors and deep learning algorithms can implement many interesting tasks. As shown in Fig.~\ref{fig:manipulation}(h), Church \textit{et al.} combined visuotactile sensors and reinforcement learning for keyboard input\cite{church2020deep}.
 Wang \textit{et al.} implemented pen flip operations using an end-to-end supervised learning channel based on tactile exploration~\cite{wang2020swingbot}. Dong \textit{et al.} implemented the insertion task via reinforcement learning and achieved a success rate of over 85\% in four different object insertion experiments~\cite{ dong2021tactile}.

\subsection{Other applications}

As an emerging technology, visuotactile perception can also be incorporated into other robotic components, such as arms~\cite{zhang2020vtacarm} and feet~\cite{stone2020walking}. Zhang \textit{et al.} proposed a smart foot that can acquire the contact surface tilt angle and foot pose using visuotactile sensing~\cite{zhang2021tactile}. In addition to the robot foot, improving the tactile perception of the robotic arm is important for improving the safety of robot interaction. Asahina \textit{et al.} proposed a robotic arm that can perceive the contact area to improve the robot's perception ability during human-robot interaction~\cite{asahina2019development}. To further improve the perception and obstacle avoidance capability of the robotic arm, Luu \textit{et al.} designed a robotic arm with controlled transparency using the PDLC(Polymer Dispersed Liquid Crystals) film, which can switch between transparent and opaque~\cite{luu2022soft}.

\section{Discussion and Perspectives}


The wide application of digital image sensors and recent leaps in computer vision boosted the development of visuotactile sensors, which enabled robots with high-resolution tactile sensation by processing image signals. However, the previously introduced prototypes are still yet to be perfect in terms of design and signal processing. We give our insights in this section for the future development of visuotactile sensors.

\subsection{Design}
Hardware and algorithms for visuotactile sensors are complementary. The hardware level improvements on the following aspects can fundamentally breakthrough the limitations and expand the applications scenarios of visuotactile sensors:
\begin{itemize}

\item Multimodality: The information modality of current visuotactile sensors is limited in visual cues, which makes them hard to accomplish complex sensing tasks. Improvements in multimodal perception capabilities can be realized by designing functional sensing layers, multi-mode illumination systems, hyperspectral image sensors, and advanced optical structures. 

\item Portability: Visuotactile sensors have the potential to provide detailed contact information, but the size of the sensors limits their development. This is because the thickness of the sensor is dominated by the focal length of the camera, which is especially difficult to shrink for 180 degrees FOV wide-angle lens. In the future, the application of optical waveguides, bio-inspired compound eyes optical structure,  CMOS technology\cite{rahiminejad2022novel}, and optical refraction technologies will further reduce the thickness of the visuotactile sensor. 

\item Flexibility: Most of the current visuotactile sensors only have the sensing skin part soft. Although some flexible robotic fingers have been proposed~\cite{liu2022GelSight,she2020exoskeleton}, flexible fingertip sensors still need further investigation. The development of flexible electronics, photonics, and material science are expected to provide solutions in achieving the overall flexibility of visuotactile sensors.

\item Sensitivity: Visuotactile sensors calculate the amount of contact force by analyzing the deformation of the sensory skin. Since small forces are difficult to deform the sensory skin, the detection of small forces is a major challenge for the visuotactile sensors. In addition to small forces, the perception of microscopic texture is also challenging. The application of super-resolution technology and microscopic imaging technology will be of great help to improve the sensitivity of the visuotactile sensor.


\end{itemize}

\subsection{Signal processing}
The quest in signal processing techniques on the following topics is expected to more thoroughly exploit the information from visuotactile sensors' output:
\begin{itemize}

\item Light field control: The mainstream method in 3D reconstruction is the photometric stereo method~\cite{yuan2017GelSight}, which requires a highly precise optical path to guarantee its accuracy. Future development with controllable structured light may bring a significant improvement in the reconstruction accuracy of the visuotactile sensor. The use of ordinary light to reconstruct the sensor surface can also drastically promote the development of visuotactile sensors.

\item Multi-sensor fusion: By combining multiple visuotactile sensors with vision, acoustic, and even chemical sensors, intelligent robots with human-like perception may achieve higher-level cognitive functions and facilitate complex manipulation tasks. Future research in compiling high-dimensional robotic perception models is an essential step to create the next generation of intelligent robots.

\item Closed-loop control frameworks: Most existing works on visuotactile sensors only focus on improving their perceptive functions. Combining visuotactile sensing technology into closed-loop control frameworks will greatly improve the operation ability of robots.

\item Tactile reconstruction and localization: Although some research has been conducted on depth reconstruction and perception, the combination of visuotactile perception and depth reconstruction algorithms is still of high technical value in solving object reconstruction under occlusion or low-visibility situations.

\item Commercialization: Although a wide variety of visuotactile sensors have been proposed, not many of them received commercial success. Future works in hardware standardization, user-friendly calibration process, unified interface for different robotic systems (e.g., The Robot Operating System (ROS)), and improvement in durability can accelerate the adoption and commercialization of visuotactile sensors.

\item Realistic simulation engine: Although the current physical simulation engines are able to simulate and realize the simulation of light, texture, and deformation in the process of contact with objects, they mainly consider the reflection, brightness, and direction of light. Future inclusion of the sensor surface material may improve the process of sim-to-real.

\item Task-orientated optimization: Currently, visuotactile sensors are mainly used in the field of object grasping for indoor scenes. In fact, visuotactile sensors with a large area and high resolution are advantageous for improving robot object grasping in low-visibility environments, such as darkness, smoke, underwater, and other extreme scenarios. In addition to grasping, exploring the usage of visuotactile sensors in industrial, emergency rescue, entertainment, medical and other scenarios could be meaningful.
m
\item Large tactile language models: Large Language Models (LLM) are becoming increasingly widely used in people's lives, and combining visuotactile perception with LLM will further enhance the robot's operation performance.

\end{itemize}

\section{Conclusion}

Visuotactile perception fully combines the advantages of high resolution in visual perception and high reliability in tactile perception, enabling perception of not only the contact positions, but also contact forces, slip information, and object pose through advanced signal processing algorithms. Despite some progress in visuotactile sensor design, issues such as thickness and hardness still limit their development. Future research can address this by integrating emerging sensing materials and technologies into the design of sensing skin, thus expanding the range of applications for visuotactile sensors. Regarding algorithms, while current models are capable of providing valuable information through signal processing, most can only accomplish one function. In the future, the development of a general-purpose large model capable of outputting multimodal information may significantly amplify the functionality of visuotactile sensors.

In a word, the field of visuotactile perception contains many unknown areas and this article reviews current technologies for visuotactile perception from the perspective of signal processing. We hope this review can give readers a more comprehensive understanding of visuotactile sensing technology from a different angle and thus further promote signal processing development in this field.

\bibliographystyle{IEEEtran}
\bibliography{ref}

\begin{IEEEbiography}[{\includegraphics[width=1in,height=1.25in,clip,keepaspectratio]{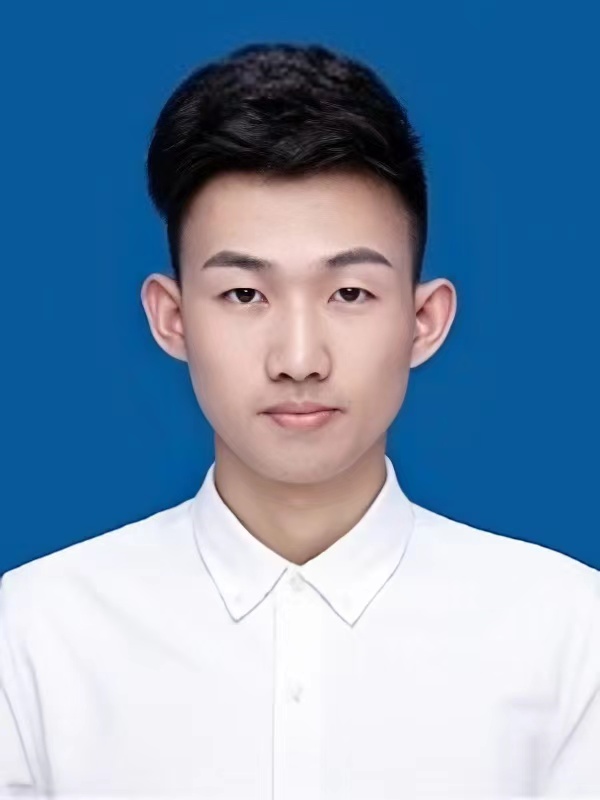}}]{Shoujie Li}
received the B.Eng. degree in electronic information engineering from the College of Oceanography and Space Informatics, China University of Petroleum,
Tsingtao, China, in 2020. He is currently pursuing toward Ph.D. degree in Tsinghua-Berkeley Shenzhen Institute, Shenzhen International Graduate School, Tsinghua University, Shenzhen, China.

His research interests include tactile perception, grasping, and machine learning. He received the Outstanding Mechanisms and Design Paper Finalists in 2022 ICRA and the Best Application Paper Finalists in 2023 IROS. He won first place in the Robotic Grasping of Manipulation Competition-Picking in Clutter in 2024 ICRA. 
\end{IEEEbiography}

\begin{IEEEbiography}[{\includegraphics[width=1in,height=1.25in,clip,keepaspectratio]{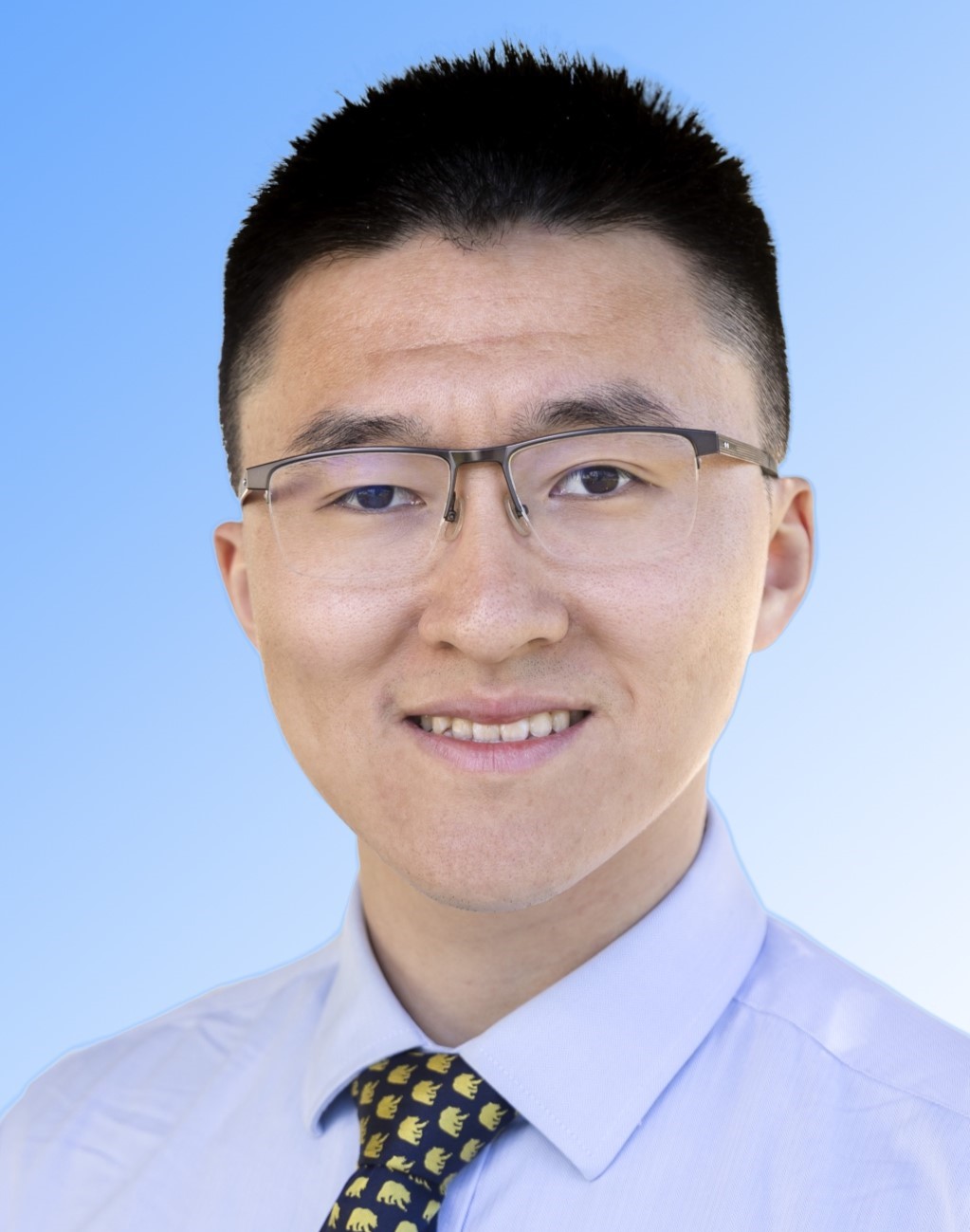}}]{Zihan Wang}
received his dual-B.Eng. Degrees from Xidian University and Heriot-Watt University (1st Class Hons.) in 2019, and his Ph.D. Degree from Tsinghua University in 2024. Currently, he is a postgraduate researcher at the University of California, Berkeley, under the supervision of Professor Liwei Lin. 

His research interests include sensors and actuators with their applications in robotics and wearable devices.
\end{IEEEbiography}

\begin{IEEEbiography}
[{\includegraphics[width=1in,height=1.25in,clip,keepaspectratio]{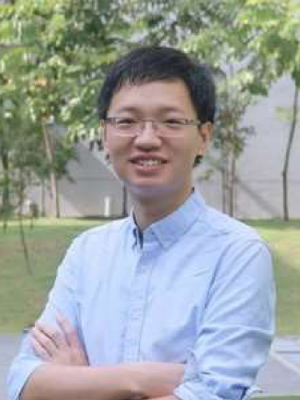}}]{Changsheng WU}
 is a Presidential Young Professor in the Department of Materials Science and Engineering (MSE) at the National University of Singapore (NUS). He is also an assistant professor by courtesy in Electrical and Computer Engineering and a PI in the Institute for Health Innovation and Technology and the N.1 Institute for Health, NUS. He received his PhD in MSE from Georgia Tech and carried out postdoctoral research in the Querrey Simpson Institute for Bioelectronics at Northwestern University.
 
 His research focuses on developing wireless wearables and intelligent robots for energy harvesting, biosensing, and therapeutic applications, leveraging bioelectronics, materials science, and advanced manufacturing to create solutions for sustainable living and the environment.
\end{IEEEbiography}

\begin{IEEEbiography}
[{\includegraphics[width=1in,height=1.25in,clip,keepaspectratio]{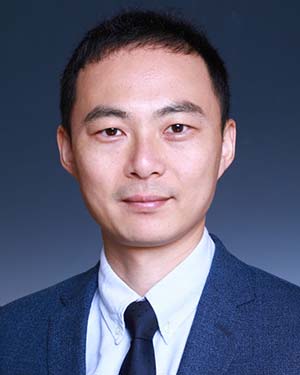}}]{Xiang Li}
 (Senior Member, IEEE) received the Doctor degree from Nanyang Technological University in 2013. He is currently an Associate Professor with the Department of Automation, Tsinghua University, Beijing, China. His research interests include robotic manipulation, micro/nano robots, and human–robot interaction. He received the Best Application Paper Finalist in 2017 IROS and the Best Medical Robotics Paper Finalist in 2024 ICRA. He led the XL team, won the first place in the 2024 ICRA Robotic Grasping and Manipulation Challenge In-Hand Manipulation Track, and also received the “Most Elegant Solution” award across all tracks. 
 
 He has been the Associate Editor of IEEE Robotics and Automation Letters since 2022 and the Associate Editor of IEEE Transactions on Automation Science and Engineering since 2023. He is the Program Chair of the 2023 IEEE International Conference on Real-time Computing and Robotics.

\end{IEEEbiography}

\begin{IEEEbiography}
[{\includegraphics[width=1in,height=1.25in,clip,keepaspectratio]{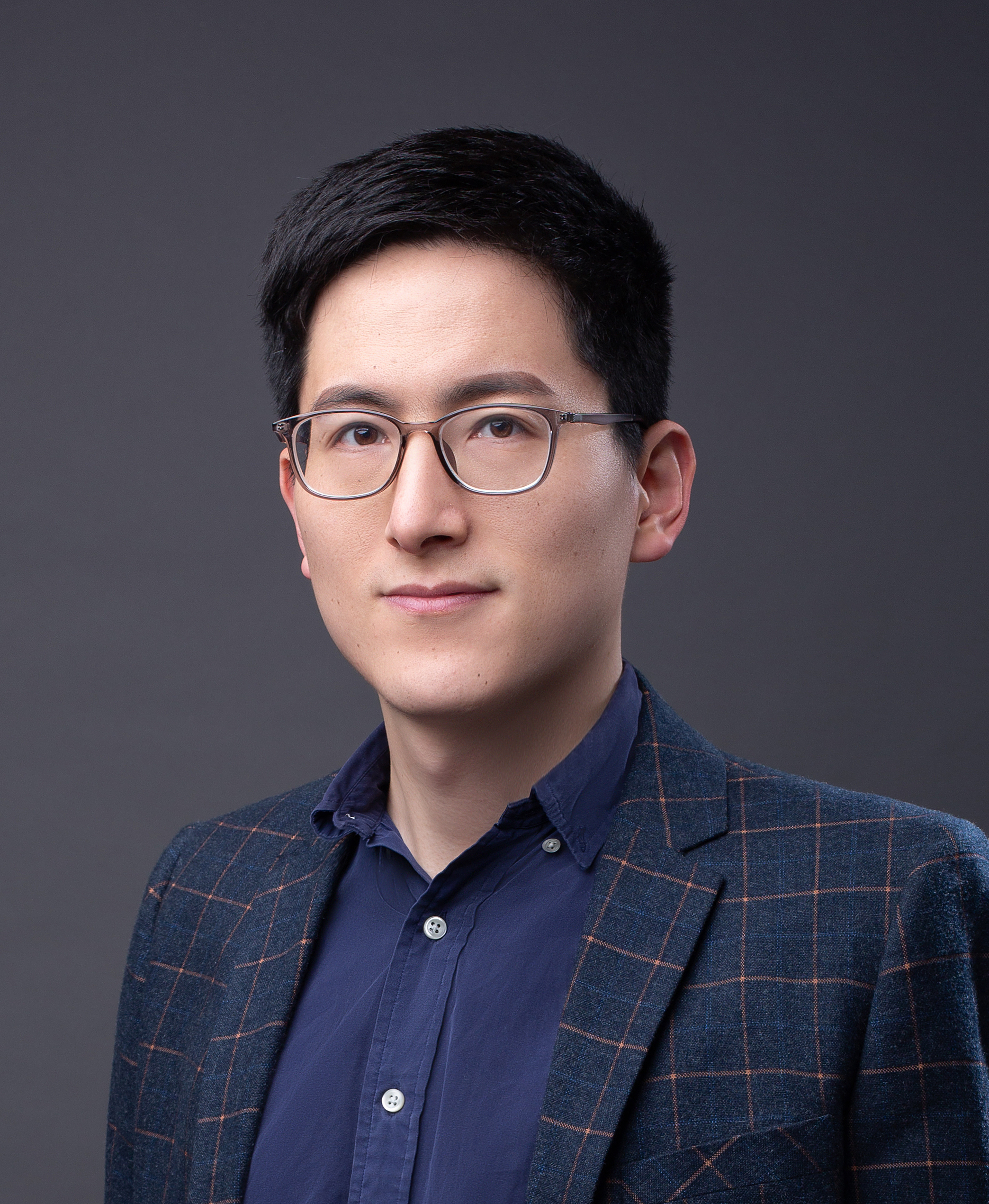}}]{Shan Luo}

Shan Luo is a Reader (Associate Professor) at the Department of Engineering, King’s College London. Previously, he was a Lecturer at the University of Liverpool, and Research Fellow at Harvard University and University of Leeds. He was also a Visiting Scientist at the Computer Science and Artificial Intelligence Laboratory (CSAIL), MIT. He received the B.Eng. degree in Automatic Control from China University of Petroleum, Qingdao, China, in 2012. He was awarded the Ph.D. degree in Robotics from King’s College London, UK, in 2016. 

His research interests include tactile sensing, robot learning and robot visual-tactile perception. He is a Senior Member of the IEEE.

\end{IEEEbiography}

\begin{IEEEbiography}
[{\includegraphics[width=1in,height=1.25in,clip,keepaspectratio]{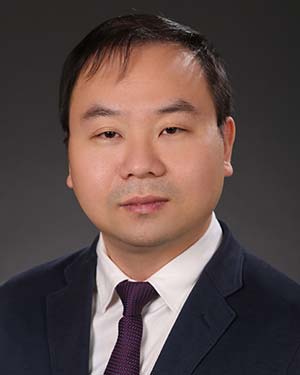}}]{Bin Fang}

 (Senior Member, IEEE) received the Ph.D. degree in mechanical engineering from Beihang University, Beijing, China, in 2014. He was a Research
Assistant with the Department of Computer Science and Technology, Tsinghua University, Beijing. He is currently a Professor with School of Artificial Intelligence, Beijing University of Posts and Telecommunications, Beijing, China.

His research interests include tactile sensors, soft robots, and human–robot interaction.

\end{IEEEbiography}

\begin{IEEEbiography}
[{\includegraphics[width=1in,height=1.25in,clip,keepaspectratio]{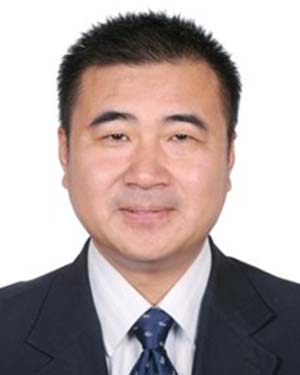}}]{Fuchun Sun}

 (S’94—M’98—SM’07—F’19) was born in Jiangsu Province, China, in 1964. He received the Ph.D degree from the Department of Computer Science and Technology, Tsinghua University, Beijing, China, in 1998. After completing his Ph.D studies on Computer Applications in Tsinghua University in March 1998. In April 2001, He joined Department of Computer Science and Technology, Tsinghua University. Now he is a full professor in the Department of Computer Science and Technology, Tsinghua University, Beijing, China.
 
His research interests include robot active sensing, cross-modal learning, and skill learning for robot manipulations. He won the Excellent Doctoral Dissertation Prize of China early in 2000 and the Choon-Gang Academic Award by Korea in 2003, and was recognized as a Distinguished Young Scholar in 2006 by the National Science Foundation of China. In the past ten years, his research work transferred to skill learning and cognitive computation of robots using vision, tactile and auditory sensing. He has authored or coauthored two books and over 300 papers which have appeared in various journals and conference proceedings. He received Andy Chi best paper award by IEEE Instrumentation \& Measurement Society in 2017. His team also won the first place in IROS 2016, 2019 and ICRA 2024 robotic grasping, assembly and sim to real competition. He was elected as IEEE Fellow in 2018, CAAI Fellow in 2019 and CAA in 2020.

He is now the EIC of the international journals of Cognitive Computation and Systems, AI and Autinumous Systems, and also serves as AEs in IEEE Trans. On Fuzzy Systems.

\end{IEEEbiography}

\begin{IEEEbiography}
[{\includegraphics[width=1in,height=1.25in,clip,keepaspectratio]{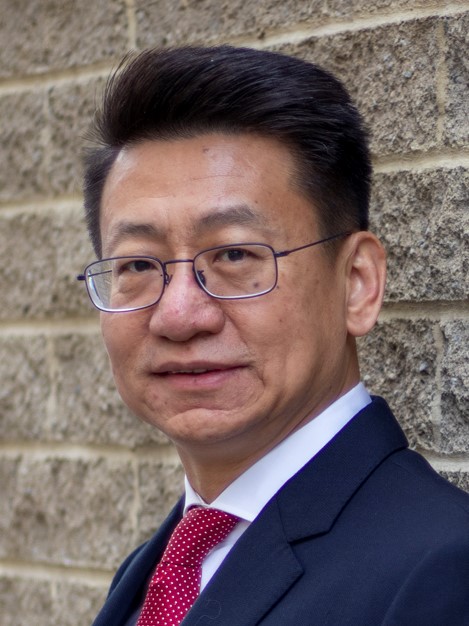}}]{Xiao-Ping Zhang}

received B.S. and Ph.D. degrees from Tsinghua University, in 1992 and 1996, respectively, both in Electronic Engineering. He holds an MBA in Finance, Economics and Entrepreneurship with Honors from the University of Chicago Booth School of Business, Chicago, IL. 

He is the founding Dean of Institute of Data and Information (iDI) at Tsinghua Shenzhen International Graduate School (SIGS), Chair Professor at Tsinghua SIGS and Tsinghua-Berkeley Shenzhen Institute (TBSI), Tsinghua University. He had been with the Department of Electrical, Computer and Biomedical Engineering, Toronto Metropolitan University (Formerly Ryerson University), Toronto, ON, Canada, as a Professor and the Director of the Communication and Signal Processing Applications Laboratory (CASPAL), and has served as the Program Director of Graduate Studies. His research interests include statistical signal processing, image and multimedia content analysis, machine learning/AI/robotics, sensor networks and IoT, and applications in big data, finance, and marketing.

Dr. Zhang is Fellow of the Canadian Academy of Engineering, Fellow of the Engineering Institute of Canada, Fellow of the IEEE, a registered Professional Engineer in Ontario, Canada, and a member of Beta Gamma Sigma Honor Society. He is the general Co-Chair for the IEEE International Conference on Acoustics, Speech, and Signal Processing, 2021. He is the general co-chair for 2017 GlobalSIP Symposium on Signal and Information Processing for Finance and Business, and the general co-chair for 2019 GlobalSIP Symposium on Signal, Information Processing and AI for Finance and Business. He was an elected Member of the ICME steering committee. He is the general chair for ICME2024. He is Editor-in-Chief for the IEEE JOURNAL OF SELECTED TOPICS IN SIGNAL PROCESSING. He is Senior Area Editor for the IEEE TRANSACTIONS ON IMAGE PROCESSING. He served as Senior Area Editor the IEEE TRANSACTIONS ON SIGNAL PROCESSING and Associate Editor for the IEEE TRANSACTIONS ON IMAGE PROCESSING, the IEEE TRANSACTIONS ON MULTIMEDIA, the IEEE TRANSACTIONS ON CIRCUITS AND SYSTEMS FOR VIDEO TECHNOLOGY, the IEEE TRANSACTIONS ON SIGNAL PROCESSING, and the IEEE SIGNAL PROCESSING LETTERS. He was selected as IEEE Distinguished Lecturer by the IEEE Signal Processing Society and by the IEEE Circuits and Systems Society.

\end{IEEEbiography}

\begin{IEEEbiography}
[{\includegraphics[width=1in,height=1.25in,clip,keepaspectratio]{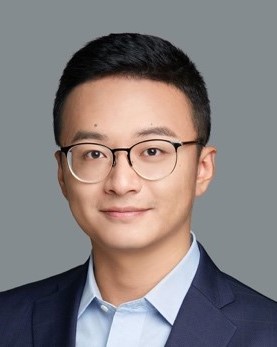}}]{Wenbo Ding}
 received the BS and PhD degrees (Hons.) from Tsinghua University in 2011 and 2016, respectively. He worked as a postdoctoral research fellow at Georgia Tech under the supervision of Professor Z. L. Wang from 2016 to 2019. He is now an associate professor and PhD supervisor at Tsinghua-Berkeley Shenzhen Institute, Institute of Data and Information, Shenzhen International Graduate School, Tsinghua University, where he leads the Smart Sensing and Robotics (SSR) group. He has received many prestigious awards, including the Gold Medal of the 47th International Exhibition of Inventions Geneva and the IEEE Scott Helt Memorial Award.
 
 His research interests are diverse and interdisciplinary, which include self-powered sensors, energy harvesting, and wearable devices for health and robotics with the help of signal processing, machine learning, and mobile computing. 
\end{IEEEbiography}

\end{document}